%% file: main.tex
\documentclass[sigconf]{acmart}

\usepackage[table]{xcolor} 
\usepackage[raster,skins, most]{tcolorbox} 
\usepackage{pifont}
\newcommand{\cmark}{\ding{51}}
\newcommand{\xmark}{\ding{55}}
\usepackage{booktabs,multirow,makecell} 
\usepackage{paralist}   
\usepackage{cleveref}
\crefname{figure}{Figure}{Figures}
\crefname{table}{Table}{Tables}
\crefname{appendix}{Appendix}{Appendixes}
\AtBeginDocument{%
  }

\setcopyright{acmlicensed}
\copyrightyear{2018}
\acmYear{2018}
\acmDOI{XXXXXXX.XXXXXXX}
\acmConference[Conference acronym 'XX]{Make sure to enter the correct
  conference title from your rights confirmation email}{June 03--05,
  2018}{Woodstock, NY}
\acmISBN{978-1-4503-XXXX-X/2018/06}

\begin{document}
\title{LLMAtKGE: Large Language Models as Explainable Attackers against Knowledge Graph Embeddings}

\author{Ting Li}
\email{liting226@mail2.sysu.edu.cn}
\orcid{0009-0009-2105-5602}
\affiliation{
  \institution{Sun Yat-sen University}
  \city{Guangzhou}
  \country{China}
}

\author{Yang Yang}
\email{yangy2233@mail2.sysu.edu.cn}
\orcid{}
\affiliation{
  \institution{Sun Yat-sen University}
  \city{Shenzhen}
  \country{China}
}

\author{Yipeng Yu}
\email{yypzju@163.com}
\orcid{}
\affiliation{
  \institution{Alibaba Inc.}
  \city{Hangzhou}
  \country{China}
}

\author{Liang Yao}
\authornote{Corresponding author.}
\email{yaoliang3@mail.sysu.edu.cn}
\orcid{}
\affiliation{
  \institution{Sun Yat-sen University}
  \city{Shenzhen}
  \country{China}
}

\author{Guoqing Chao}
\email{guoqingchao@hit.edu.cn}
\orcid{}
\affiliation{
  \institution{Harbin Institute of Technology}
  \city{Weihai}
  \country{China}
}

\author{Ruifeng Xu}
\email{xuruifeng@hit.edu.cn}
\orcid{}
\affiliation{
  \institution{Harbin Institute of Technology}
  \city{Shenzhen}
  \country{China}
}

\renewcommand{\shortauthors}{Ting Li et al.}

\begin{abstract}
Adversarial attacks on knowledge graph embeddings (KGE) aim to disrupt the model's ability of link prediction by removing or inserting triples. A recent black-box method has attempted to incorporate textual and structural information to enhance attack performance. However, it is unable to generate human-readable explanations, and exhibits poor generalizability. In the past few years, large language models (LLMs) have demonstrated powerful capabilities in text comprehension, generation, and reasoning. In this paper, we propose LLMAtKGE, a novel LLM-based framework that selects attack targets and generates human-readable explanations. To provide the LLM with sufficient factual context under limited input constraints, we design a structured prompting scheme that explicitly formulates the attack as multiple-choice questions while incorporating KG factual evidence. To address the context-window limitation and hesitation issues, we introduce semantics-based and centrality-based filters, which compress the candidate set while preserving high recall of attack-relevant information. Furthermore, to efficiently integrate both semantic and structural information into the filter, we precompute high-order adjacency and fine-tune the LLM with a triple classification task to enhance filtering performance. 
Experiments on two widely used knowledge graph datasets demonstrate that our attack outperforms the strongest black-box baselines and provides explanations via reasoning, and showing competitive performance compared with white-box methods. Comprehensive ablation and case studies further validate its capability to generate explanations. 
\end{abstract}




\maketitle

\begin{figure}[!htbp]
    \centering
    \includegraphics[width=1\linewidth]{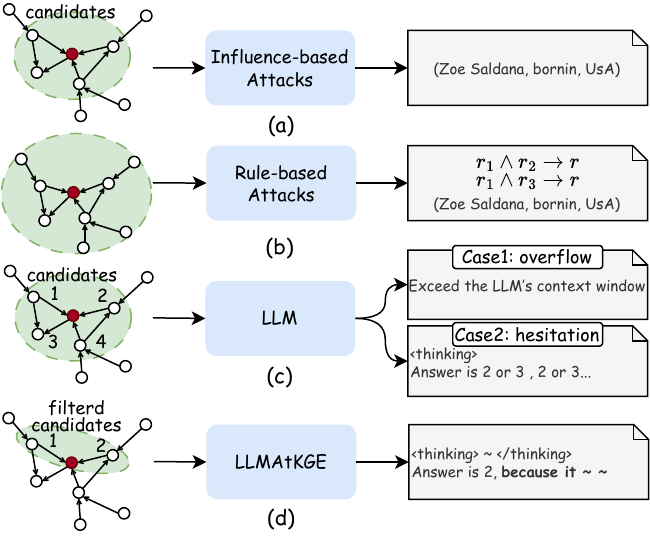}
    \caption{Comparison of current methods and our proposed LLMAtKGE. (a) Influence-based attacks typically rely on 1-hop triples as input and lack the ability to generate explanations. (b) Rule-based attacks exploit rules extracted from the KG to perform attacks, and derive explanations from selected rules. (c) An LLM-based attempt, even though providing only 1-hop triples may still lead to overflow, and hesitation in the chain of thought. (d) Our LLMAtKGE guides the LLM's reasoning to derive the answer while generating the human-readable explanations.}
    \label{fig:intro}
    \Description{}
\end{figure}

\section{Introduction}
Knowledge graph (KG) is widely used to model structured knowledge and relationships. Knowledge graph embedding (KGE) is a representation learning method that maps entities and relations in a knowledge graph into a low-dimensional vector space. In recent years, KGE has been successfully applied to various downstream tasks, such as knowledge graph completion (KGC)~\cite{TransE,sun2019rotate}, knowledge graph question answering (KGQA)~\cite{saxena2020improving,huang2019knowledge}, and semantic search~\cite{gerritse2020graph,nikolaev2020joint}. 
Existing studies~\cite{CRIAGE,bhardwaj2021poisoning} have demonstrated that KGEs are vulnerable to attacks. Some black-box methods~\cite{BaseAttack,PLM-Attack} have demonstrated that the prediction of a target triple can be disrupted without accessing KGE's gradients, simply by adding or removing a single triple corresponding to each test triple in the training set. 

As shown in~\cref{fig:intro}, we roughly group these black-box attack methods into two categories: (1) influence-based approaches~\cite{BaseAttack,PLM-Attack} identify the most influential triples as attack targets by calculating the similarity among triple embeddings; (2) rule-based approaches~\cite{AnyBURL,zhao2024untargeted} generate attack targets by capturing patterns and learning logic rules from the KG's structure. 
Although some of these studies claim that their approaches are explainable attacks, they fail to provide human-readable explanations that clarify why the attacks succeed and how the robustness of KGE models can be improved.
Fortunately, large language models (LLMs) have demonstrated impressive performance in a range of KG tasks~\cite{chen2025knowledge, yao2025exploring}, thanks to their powerful generation ability. By leveraging this strength, LLMs have great potential to produce human-readable explanations, making them more practical and effective attackers. 
Thus, we pay attention to \textbf{RQ1: how to use LLMs as explainable attackers against KGEs?}
In practice, KGE attacks usually rely on heuristic methods~\cite{zhang2019data, bhardwaj2021poisoning}, large-scale enumeration to discover effective perturbations, and the lack of labeled datasets makes it difficult to perform supervised fine-tuning of LLMs. 
To address this, we propose a parameter-preservation framework that enhances LLM with KG to generate attack targets, as shown in~\cref{fig:intro}. Specifically, the target triple and its description are used as KG factual evidence, the neighboring triples of it serve as the candidates. They are concatenated into a context and fed into the instruction LLM, which reasons to select a perturbation target and generate human-readable explanations. 

Ideally, the candidate set would include all triples of KG, but the context window of LLM is limited. 
Moreover, compared with question answering task based on text retrieval~\cite{wang2024learning,luo2024reasoning,gutierrez2024hipporag,edge2025local}, KGE attacks require cross-relation structural reasoning. An oversized candidate set enlarges the search space and induces hesitation in the chain of thought. Specifically, the LLM hesitates among multiple perturbation targets, which prolongs reasoning and may exceed the token budget. 
Therefore, we introduce semantics-based and centrality-based filters, which narrow the candidate set from the perspectives of semantics and graph structure while retaining attack-relevant information as much as possible. 

However, since semantics-based and centrality-based filters do not jointly capture both semantics and structural information, their performance remains limited. Moreover, complex relations and hierarchical structures are difficult to convey clearly through context, restricting the LLM's view. Thus, we focus on \textbf{RQ2: how to simply and effectively integrate multi-hop information into LLM?} 
To address these challenges and enhance filters, we explore that seamlessly integrate the KG with the LLM through parameter updates. Specifically, we align the LLM with the KG via a triple-level task, thereby broadening the view on KG and improving its ability of filtering. 
Existing methods~\cite{KoPA} leverage lightweight KGE to fine-tune LLMs for approximating structural features, rather than explicitly incorporating multi-hop paths, which causes the LLM to overemphasize semantic similarity and ultimately limits performance. For example, the embedding of "20th United States Congress" is closest to "21st United States Congress" instead of the members of the 20th Congress. 
The key challenge is to balance the contributions of semantic and structural signals, efficiently embed multi-hop path information, and meanwhile preserve the LLM's existing knowledge. Inspired by classical GNNs~\cite{hamilton2017inductive,kipf2017semisupervised,velickovic2018graph}, we leverage neighborhood aggregation, which exhibits strong structure-aware capabilities. It encodes multi-hop information via the interaction of the adjacency and feature matrices. A knowledge graph is inherently a graph structure. We propose an efficient adapter that precomputes high-order adjacency (HoA) to transform multi-hop KG information into the textual embedding space, serving as prefix tokens for LLM tuning.

We conduct deletion and addition attacks on four KGE models, including DistMult~\cite{DistMult}, ComplEx~\cite{ComplEx}, ConvE~\cite{WN18RR}, and TransE~\cite{TransE}, using two widely used knowledge graph datasets, WN18RR and FB15k-237. Our proposed attack method outperforms the strongest black-box baseline in most cases, and even surpasses white-box attacks to achieve state-of-the-art performance. We further perform extensive ablation studies to validate the effectiveness of each component of our framework. Finally, a case study demonstrates the ability of our model to generate human-readable explanations. With few open releases in recent work, \textbf{our code\footnote{\url{https://github.com/liting1024/LLMAtKGE}} is publicly available.} 

Our contributions are summarized in three aspects. 
\begin{itemize}
    \item We design a novel parameter-preservation framework called \textbf{\underline{LLMAt}}tack\textbf{\underline{KGE}}, which enables LLMs to perform KGE attacks in the absence of labeled data. It employs prompt engineering to generate human-readable explanations, while introducing filters to address overflow and the hesitation issue. 
    \item We introduce a paradigm that precomputes high-order adjacency, then transforms multi-hop information into the LLM embedding space. It enables the LLM to capture deeper structural knowledge and improves the filtering ability. 
    \item We conduct extensive experiments on two widely used knowledge graph datasets and demonstrate the effectiveness of our method, and design ablation experiments to rigorously evaluate the contribution of each module. 
\end{itemize}

\section{Related Work}
\subsection{Bridging KGs and LLMs}
With the rapid advancement of LLMs, research on their integration with KG can be broadly divided into two categories by task type and integration strategy. 
KG-enhanced LLM primarily focus on semantic retrieval, including KGQA and document QA, with recent studies exploring improvements through in-context learning (ICL) and retrieval-augmented generation (RAG). In contrast, LLM-enhanced KG emphasize reasoning, and recent works aim to integrate structural information to strengthen inference capabilities. 
As our task requires reasoning on KGs but suffers from limited labeled datasets, directly fine-tuning an LLM to combine KG and textual information is infeasible. Therefore, we combine the benefits of KG-enhanced LLMs and LLM-enhanced KGs, employing ICL for knowledge alignment and then introducing triple classification task to embed structural information into the LLM. 

\subsubsection{KG-enhanced LLM}
This category leverage knowledge graphs to mitigate hallucinations, provide reasoning paths, and enhance both interpretability and generation quality. 
For example, GcR~\cite{luo2025graphconstrained} introduces a KG Trie structure as a constraint, enabling LLMs to follow multiple reasoning paths and thus achieve more faithful multi-path reasoning. EtD~\cite{liu2025dual} employs a GNN to adaptively expand and prune the KG according to the question semantics, identifying candidate entities and evidence chains. RoG~\cite{luo2024reasoning} enhances interpretability by avoiding direct SPARQL generation, the LLM first produces relation paths as reasoning plans, which are then validated through KG path retrieval to generate answers with explicit evidence. ToG~\cite{sun2024thinkongraph} also performs entity and relation search, pruning, and beam-search reasoning over the KG to construct traceable reasoning paths. Building on this idea, ToG2~\cite{ma2025thinkongraph} further integrates textual RAG with KG retrieval to enable more faithful reasoning. 
Knowledge graphs have also been leveraged to enhance document QA. GraphRAG~\cite{guo2024lightrag,sarthi2024raptor} extends RAG by exploiting KG semantics and relational paths for document retrieval and reasoning. Edge et al.~\cite{edge2025local} builds an KG with community-level summaries to support global corpus understanding. HippoRAG~\cite{gutierrez2024hipporag,gutierrez2025rag}, inspired by hippocampal memory indexing theory, combines KG and personalized PageRank with context augmentation and memory recognition to improve global comprehension and multi-hop retrieval.

\subsubsection{LLM-enhanced KG}
LLMs have also been employed for reasoning over KG to support construction and completion, including triple classification, entity prediction, and relation prediction. 
KG-LLM~\cite{yao2025exploring} serializes entities, relations, and triples into natural language QA forms, and employs instruction tuning to enable LLMs to perform KGC tasks. MKGL~\cite{guo2024mkgl} introduces specialized KG language tokens to represent triples, and integrates word embeddings with a principal neighbourhood aggregator for reasoning, thereby achieving ranking prediction over the entire entity set. K-ON~\cite{guo2025kon} converts KG into token-level inputs and adds a multi-head prediction layer on top of the LLM, enabling entity-level contrastive learning and efficient knowledge alignment with only modifications to the output layer. 
KICGPT~\cite{wei2023kicgpt} leverages the vast commonsense knowledge encoded in pretrained LLMs as an external source, combining structure-aware KG models with LLMs and applying in-context learning to guide re-ranking over Top-k results. MPIKGC~\cite{xu2024multiperspective} further proposes a multi-perspective augmentation strategy, where LLMs enrich textual descriptions of KGs from entity, relation, and structural perspectives, and the augmented knowledge is then fed into description-driven KGC models. 

\subsection{Adversarial Attacks on KGEs}
Adversarial attacks are typically conducted by poisoning the KGE parameters or their datasets. 
Early studies mainly focused on white-box or gray-box attack settings, which require access to model parameters. 
CRIAGE~\cite{CRIAGE} is limited to attacking multiplicative models, it leverages the first-order Taylor to approximate embedding variations and trains an inverter network to narrow the search space. Furthermore, Bhardwaj et al.~\cite{BaseAttack} introduced gradient similarity and influence Function, achieving attacks on multiplicative and additive models by utilizing model gradients. 
In both of their works, Bhardwaj et al.~\cite{bhardwaj2021poisoning} use embedding similarity to guide parameter search in attacks. 
Recently, Kapoor et al.~\cite{kapoor2025robustness} introduced graph, label, and parameter perturbation to perform untargeted attacks through entity or relation replacement, label perturbation, and direct parameter modification.

Another category is black-box attacks. PLM-Attack~\cite{PLM-Attack} encodes triple semantics with a pretrained language model and integrates graph structure via random walks to identify high-impact triples for attacks. AnyBURL~\cite{AnyBURL} learns rules and applies abductive reasoning to generate explanations. Zhao et al.~\cite{zhao2024untargeted} achieve untargeted attacks by extracting rules to disrupt global KG semantics. Zhou et al.~\cite{zhou2024poisoning} propose server-initiate and client-initiate poisoning attack for federated KGE through dynamic shadow model optimization. 

In summary, white-box attack methods generally exploit gradient information from the target model as the attack basis, some black-box methods derive rule-based rationales. These methods still struggle to provide human-readable explanations. 

\section{Preliminary}
\subsection{Problem Definition}
\subsubsection{Knowledge Graph}
A knowledge graph can be formally defined as $(\mathcal{E},\mathcal{R},\mathcal{T})$, where $\mathcal{E}$ and $\mathcal{R}$ denote the entity set and relation set, respectively. Each element in these sets is associated with a textual description. The triple set $\mathcal{T}$ consists of triples $t=(s,r,o)$, where each triple is composed of a subject, a relation, and an object. 
Essentially, the knowledge graph can be represented as a directed graph $\mathcal{G}_\text{KG}=(\mathcal{E},\mathbf{A})$, where $\mathcal{E}$ denotes the set of nodes corresponding to entities in the KG, and $\mathbf{A}$ is the adjacency matrix.

\subsubsection{Triple Graph}
The triple graph is defined as $\mathcal{G}_\text{TG}=(\mathcal{V},\mathcal{L})$, where $\mathcal{V}$ denotes the set of triple nodes corresponding to the triple set $\mathcal{T}$ in the KG. An edge $(t_i, t_j) \in \mathcal{L}$ exists if the two triples share a common entity $e \in \mathcal{E}$. 
In addition, from the perspective of graph structures, the $h$-hop neighbors of $\mathcal{G}_\text{KG}$ and $\mathcal{G}_\text{TG}$ can be defined as $\mathcal{N}^{(h)}(v)$. 

\begin{figure*}[!ht]
    \centering
    \includegraphics[width=1\linewidth]{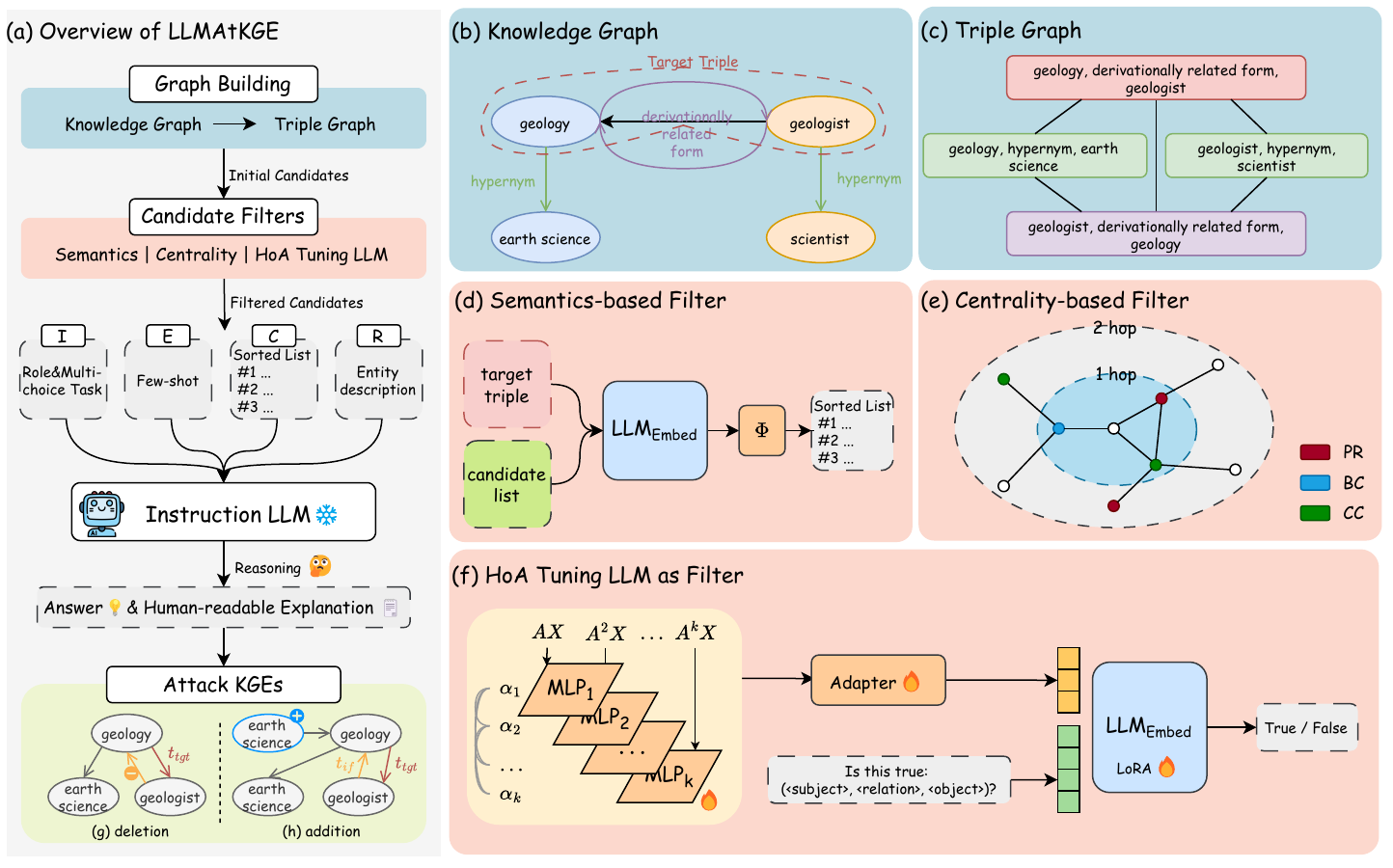}
    \caption{
    (a) The overview of LLMAtKGE. First, (b) the knowledge graph and (c) the triple graph are constructed to determine the initial candidates. The (d) semantics-based and (e) centrality-based filters are applied to shrink the candidate set. Then, following advanced prompt engineering techniques, the instruction $I$, example $E$, candidates $C$, and reference $R$ are concatenated as the input to the instruction LLM. The LLM reasons to generate answers and human-readable explanations, after which (g) deletion and (h) addition attacks are performed by removing triples or inserting new triples through entity replacement.
    (f) HoA efficiently tunes the LLM on triple classification, integrating multi-hop paths into the LLM to enhance the filtering performance. 
    }
    \label{fig:overview}
    \Description{}
\end{figure*}

\subsection{Attack on KGEs}
A KGE model $\mathcal{M}$ is trained on $\mathcal{T}_{\text{train}}$ to learn low-dimensional embeddings of entities and relations, and is evaluated on $\mathcal{T}_{\text{train}}$ via link prediction.  
The attack objective is to maximally degrade $\mathcal{M}$'s link prediction performance on targeted triples, as measured by common ranking metrics, including Mean Reciprocal Rank (MRR) and Hits@k. 

We consider the following four classical models as the attacked KGE. 
\textbf{TransE}~\cite{TransE} is a distance-based additive model that treats relations as translation vectors, enforcing $s + r \approx o$. 
\textbf{DistMult}~\cite{DistMult} is a diagonal bilinear model using element-wise multiplication $s^\top \mathrm{diag}(r) o$ . 
\textbf{ComplEx}~\cite{ComplEx} is a bilinear model in the complex space, where conjugation introduces phase to model asymmetric relations. 
\textbf{ConvE}~\cite{WN18RR} is a convolutional model that reshapes $s$ and $r$ into 2D, applies convolution to capture nonlinear interactions, and matches with $o$. 

\subsection{KG-driven LLMs}
\subsubsection{Parameter Preservation}
Given a question $Q$. The first stage retrieves relevant knowledge from the KG through extractor such as $\operatorname{get\_entity/relation/path}$. Then the retrieved factual knowledge $F$ is exploited to guide the LLM in generating the answer $A$. 
$$ F=\operatorname{Extractor}(Q \mid \text{KG}), \quad A=\operatorname{LLM}(Q \oplus F). $$

\subsubsection{Parameter Updating}
The framework exploits annotated knowledge graph data to construct alignment tasks for supervised fine-tuning, such as low-rank adaptation (LoRA), prefix tuning. The structured knowledge can be seamlessly incorporated into the LLM. It is formulated as 
$\operatorname{FT}\big(\operatorname{LLM}(t;\theta), \phi \big), $
where $\theta$ is the fixed parameters, while $\phi$ represents the trainable parameters. 

\section{Methodology}
We follow the black-box attack setting~\cite{PLM-Attack}, aiming to attack $\mathcal{M}$ with a minimal perturbation budget by deleting or adding a single triple. Moreover, to reduce randomness, attack targets are limited to test triples $t_\text{tgt}$  achieving Hits@1 under all four KGEs. 
For the deletion attack, given each test triple $t_\text{tgt} = (s_\text{tgt}, r_\text{tgt}, o_\text{tgt})$, we identify the most influential triple $t_\text{if} = (s_\text{if}, r_\text{if}, o_\text{if})$ in the training set and remove it, thereby weakening the structural and semantic support for $t_\text{tgt}$.
For the addition attack, we aim to generate a single triple as the poison that maximally perturbs $t_\text{tgt}$, misleading $\mathcal{M}$. Specifically, we select an noise entity $e_\text{noise}$ from the KG and replace either $s_\text{if}$ or $o_\text{if}$, constructing a new triple $t_\text{poison} = (s_\text{if}, r_\text{if}, e_\text{noise})$ or $t_\text{poison} = (e_\text{noise}, r_\text{if}, o_\text{if})$, which is then injected into the training set. 

\subsection{Instruction Tuning LLM for KGE Attack}
\subsubsection{Prompt Engineering}
For a given target triple $t_\text{tgt}$, we select either a candidate triple $t_\text{if}$ or an entity $e_\text{noise}$ from the candidate set to conduct the aforementioned attack. This task can be formulated as a reasoning-reflection problem. 
We input the prompt as contextual guidance to steer the LLM in rationally analyzing each candidate and providing sufficient explanation for executing the attack. The prompt can be decomposed into four components, as shown in the following formulation. Specific examples are provided in~\cref{app:prompt_example}. 

$$ P= I \oplus E \oplus C \oplus R, $$
where $\oplus$ denotes text concatenation.  

$I$ is the attack instruction, which defines the role and specifies the task. We design $I$ as a multiple-choice question, explicitly requiring the LLM to select an option and provide its reasoning. This formulation effectively projects open-ended generation onto a finite decision space, thereby constraining reasoning divergence, and alleviating hallucination. 
$C$ is the candidate set $\mathcal{C}$ formatted as a list of options. 
Drawing on retrieval-augmented methods, $R$ provides factual knowledge to the LLM by serving as entity descriptions, thereby mitigating hallucinations. 
$E$ denotes an example, which is used to explicitly map inputs to outputs. In a label-limited and data-limited scenario, by leveraging the few-shot learning capability of LLMs~\cite{brown2020language}, the model can utilize these examples to activate relevant knowledge and achieve better task alignment.  

\subsubsection{Candidate Set Filtering}
Ideally, the candidate set $\mathcal{C}$ should contain all triples or entities in the KG. However, it is infeasible to provide them all as context to the LLM. Even when following existing methods~\cite{PLM-Attack}, which select only the one hop neighborhood of $t_\text{tgt}$ as the candidate set, the search space remains prohibitively large on dense datasets such as FB15k-237. Therefore, it is necessary to design a filter that further reduces the candidate set to a size $\delta$ acceptable for the LLM's context window, while ensuring that $t_{if}$ or $e_{noise}$ is still included in the candidate set. Two specialized filters are devised for deletion and addition attacks, respectively, and are implemented by applying them on $\mathcal{G}_{\text{KG}}$ and $\mathcal{G}_{\text{TG}}$. 

\paragraph{Semantics-based type}
In the KG, semantically closer triples are more likely to serve as supportive evidence, we employ semantic similarity to filter candidate triples. The 1-hop neighbors of $t_\text{tgt}$ on the $\mathcal{G}_\text{TG}$ are closely connected to it and thus exert a stronger influence. 

$$ \mathcal{C} = \operatorname{TopK} \left\{ \Phi\!\left( \operatorname{LLM}_{\text{Embed}}(t_\text{tgt}), \operatorname{LLM}_{\text{Embed}}(c) \right) \,\middle|\, c \in \mathcal{N}^{(1)}_{\text{TG}}(t_\text{tgt}) \right\}, $$
where $\operatorname{LLM}_{\text{Embed}}$ is used to map triples into high-dimensional vectors of fixed length, $\Phi$ denotes the cosine similarity.

\paragraph{Centrality-based type}
For addition attacks that target graph structure, we select noisy entities from the target's related entities to maximally mislead the model $\mathcal{M}$. In detail, we construct the candidate set by extracting the $h$-hop subgraph of $t_{\text{tgt}}$ from the KG, and choose $h=3$ to balance perturbation relevance and misleading noise, based on the KG’s diameter analysis. 
We then rank candidate entities using three heuristic centrality measures from network science and select the top candidates for injection. 
\begin{inparaenum}
    \item Pagerank (PR) is a damped random-walk–based measure, employed to simulate multi-hop reasoning. 
    \item Betweenness centrality (BC) is a shortest-path–based measure, identifying bridge nodes that frequently lie on paths between other nodes. 
    \item Closeness centrality (CC) is a distance-based measure, defined as the reciprocal of the average shortest-path length from a node to others. 
\end{inparaenum}

$$
\mathcal{C}
= \bigcup_{\varphi \in \{\operatorname{PR},\operatorname{BC},\operatorname{CC}\}}
\operatorname{TopK}\left\{ \varphi(c) \,\middle|\, c\in \mathcal{N}^{(3)}_{\text{KG}}(s,o) \setminus \mathcal{N}^{(1)}_{\text{KG}}(s,o) \right\}.
$$

\subsection{High-order Adjacency Tuning LLM for Filtering}
Despite applying semantics-based and centrality-based filters to reduce the candidate set, the gap between semantic and structural information remains. It prevents optimal filtering effectiveness. 
Inspired by the text reranking model~\cite{zhang2025qwen3}, we introduce semantic information of KG by designing a triple classification task. Specifically, following the KG-LLM~\cite{yao2025exploring}, each triple is transformed into a natural language statement of the form “Is this true: $s \oplus r \oplus o$ ?”, with the label annotated as “True” or “False”. Then we fine-tune $\operatorname{LLM}_{\text{Embed}}$ using LoRA to enable the embedding model to embed knowledge graph triples into a high-dimensional semantic space. 
For more efficient integration of structural information into $\operatorname{LLM}_{\text{Embed}}$, we propose an adapter that leverages high-order adjacency matrices to explicitly encode structural signals. 

SGC~\cite{wu2019simplifying} simplifies multi-layer graph convolutional networks into the pre-computed feature propagation and a linear classifier, preserving the core operation while significantly reducing model complexity without compromising performance. 
Building on this idea, we first normalize $\mathbf{A}$, and then efficiently compute its $h$-th power through preprocessing, thereby enabling high-order propagation in the graph. 
Furthermore, we propose a HoA-based adapter that employs attention coefficients to adaptively fuse multi-order structural features, and subsequently injects the refined representation into the LLM.

$$ 
\begin{aligned}
\tilde{\mathbf{A}} = D^{-\frac{1}{2}} \mathbf{A} D^{-\frac{1}{2}},\quad 
H^{(k)} \leftarrow \tilde{\mathbf{A}} H^{(k-1)}, 
\end{aligned}
$$
where $\mathbf{A}$ includes self-loops and $D$ is the degree matrix. 
$H^{(0)}$ is initialized with the entity representations $\mathbf{Z}(\cdot)$ from TransE.
This process contains no trainable parameters. 

Multiple MLPs are employed to extract feature patterns from each $H^{(k)}$. A set of learnable coefficients is introduced to model the attention distribution over multi-hop representations. They are concatenated to jointly aggregate local and long-range dependencies, thereby enlarging the receptive field.

$$
\begin{aligned}
f_{\text{HoA}} &= \mathrm{MLP}\!\left( \big\|_{k=1}^{h}\, \alpha_k \cdot \mathrm{MLP}_k\!\big(H^{(k)}\big) \right),\\
\alpha_k       &= \frac{\exp(\theta_k)}{\sum_{j=1}^{h} \exp(\theta_j)} \, ,
\end{aligned}
$$
where $\theta$ denotes the learnable parameters.

\begin{table*}[!ht] 
    \caption{
    Overall results of deletion and addition adversarial attacks on WN18RR and FB15k-237. The baselines are grouped into black-box and white-box attacks. 
    Red marks the best among white-box and black-box attacks, bold black the best within black-box, underline denotes performance competitive to PLM-Attack. 
    MRR $\downarrow$ and Hits@1 $\downarrow$ indicate better attack performance. 
    }
    \label{tab:main_result}
    \centering
    \setlength{\tabcolsep}{3.8pt}      
    \renewcommand{\arraystretch}{1.25}  
    \input{tables/main_result}

\end{table*}

The adapter fuses multi-hop entity and relation representations, enabling parameter-efficient fine-tuning to inject structured knowledge into the LLM. 
$$ K=\operatorname{MLP}_\text{Adapter} \left( f_\text{HoA}(s) \| \mathbf{Z}(r) \| f_\text{HoA}(o) \right), $$
where $K$ is concatenated as a prefix to the input token sequence.


\section{Experiments}
\subsection{Experimental Setup}

We prioritize comparisons with methods that support both deletion and addition attacks and are applicable across all KGE models. \textbf{Random} attack conducts deletion by randomly removing triples, and addition by randomly replacing entities. 
Our primary baseline is \textbf{PLM-Attack}~\cite{PLM-Attack}, which adopts the same black-box attack setting and is considered the strongest competitor as it is also based on a language model. We also compare against the rule-based method \textbf{AnyBURL}~\cite{AnyBURL}. Zhao et al.~\cite{zhao2024untargeted} is not included in comparison, as it is not publicly available and uses an untargeted attack setting that is incompatible with our experimental setup.
For white-box attacks, we evaluate a simple \textbf{direct} attack~\cite{zhang2019data}, modifying facts directly  related to the target triple to change entity embeddings. And the widely recognized influence functions (\textbf{IF}) attack~\cite{BaseAttack}, which estimates the effect by approximating parameter changes to perform the attack. 

The dataset statistics, detailed settings and hyperparameter sensitivity analysis are provided in~\cref{app:settings}.

\subsection{Main Results}
\cref{tab:main_result} reports the results of deletion and addition adversarial attacks. 
Our proposed method outperforms the strongest competing PLM in most cases, even achieves the best results among black-box attacks, and in many cases surpasses white-box attacks to attain the overall best performance. Although it does not reach the state of the art in all settings, our method is more practical as it not only provides effective attack selection but also generates human-readable explanations. 
Although the performance of addition attacks has improved, this is largely attributable to our proposed framework, which can identify deceptive entities. However, the overall effectiveness still lags behind that of deletion attacks. This gap is consistent with observations reported in other baseline methods. 
Compared with PLM-Attack, the improved performance on the additive model TransE can be attributed to the combination of parameter preservation and updating, which overcomes the limitations of directly employing the Transformer architecture as an encoder. 

We further compare the attack performance across different datasets and observe that the results on WN18RR are significantly better than those on FB15k-237. We attribute this to the fact that FB15k-237 exhibits a much richer neighborhood structure, which provides stronger structural evidence for KGE learning. As a result, under the same training settings, KGE models trained on FB15k-237 generally show greater robustness, and LLMs exhibit limited generalizability to open-domain knowledge graphs. In addition, entities and descriptions in WN18RR contain abundant semantic information, much of which has already been absorbed during LLM pre-training as lexical knowledge. 

\subsection{Ablation Analysis}

\subsubsection{Impact of Prompt Settings}

\begin{table}[!htbp]
    \caption{Ablation results on prompt components examples $E$ and references $R$ for deletion attack on WN18RR.}
    \label{tab:abl_PE}
    \centering
    \renewcommand{\arraystretch}{1.2} 

\input{tables/abl_PE}

\end{table}

The prompt template comprises four components, as outlined earlier, with instruction and candidate being essential for the execution of LLM-based attacks. Therefore, we only conduct an ablation analysis to examine the contributions of examples and references. 
As shown in \cref{tab:abl_PE}, examples play a critical role in performance improvement. In the absence of $R$, introducing $E$ leads to an average increase of 17.58\% in MRR and Hits@1, with the exception of TransE. 
The contribution of $R$ on WN18RR is relatively limited, which may be attributed to the fact that the entity descriptions of WN18RR have likely been incorporated into the LLM during pre-training, and R has already been utilized when filtering the candidate set. 

\subsubsection{Comparion of Filters}

\begin{figure}[!htbp]
    \centering
    \includegraphics[width=1\linewidth]{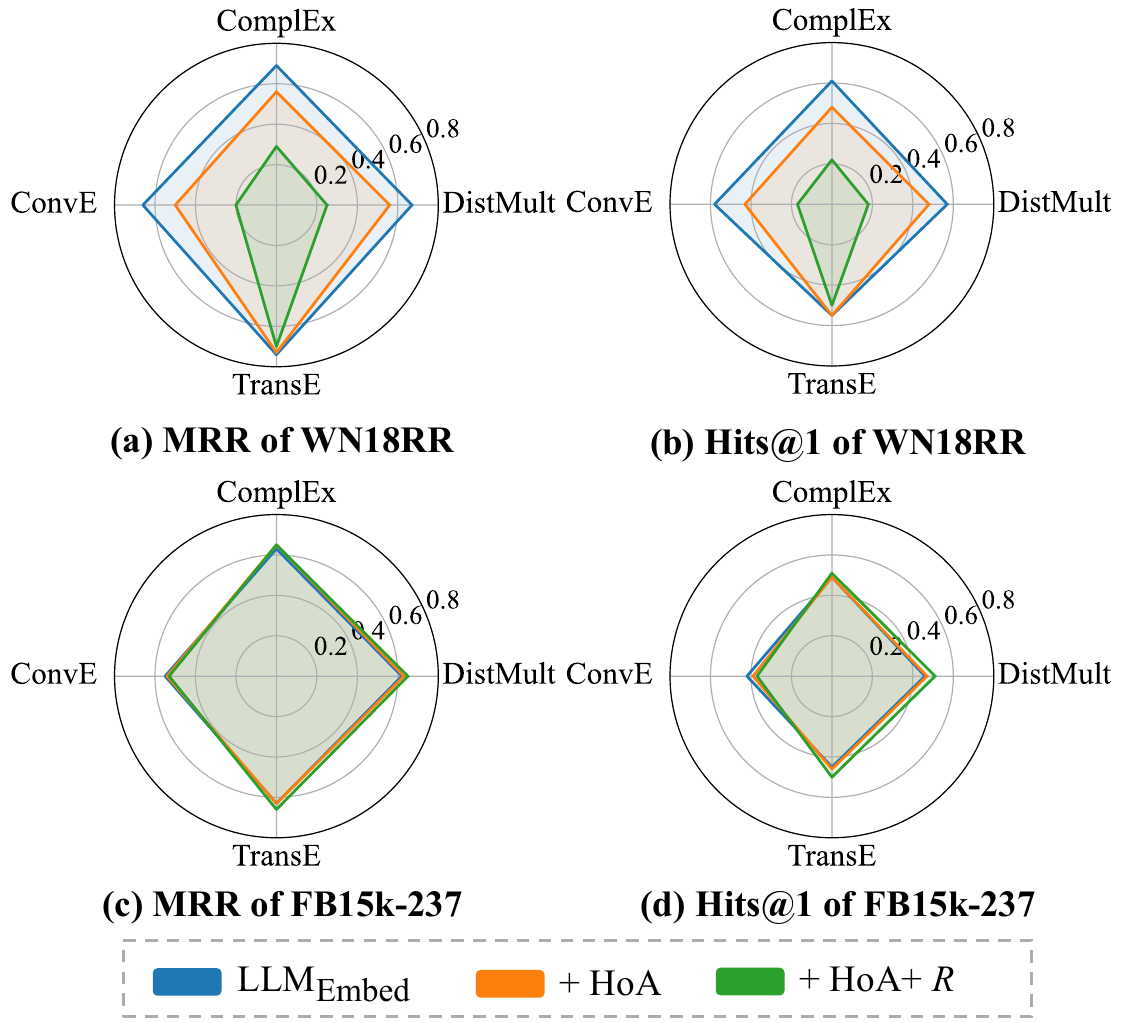}
    \caption{
    Ablation results of HoA. 
    Values closer to the center denote better attack performance. }
    \label{fig:abl_HoA}
    \Description{}
\end{figure}

To investigate the filtering effect on $r_{if}$ and $e_{noise}$, we select only the Top1 candidate as the answer. 
We adopt Llama3-8B\footnote{\url{https://huggingface.co/meta-llama/Llama-3.1-8B}} as the semantics-based filter. Although Qwen-Embedding-8B\footnote{\url{https://huggingface.co/Qwen/Qwen3-Embedding-8B}} serves as a strong competitor to Llama3-8B, its performance in the HoA fine-tuning stage is inferior, thus it is not included in the comparison. As shown in~\cref{fig:abl_HoA}, we first embed only triples to examine the effect of HoA tuning filter. HoA is effective in most cases, with particularly strong performance on WN18RR. 
Furthermore, we also embed the R of prompt to enhance the discriminative capacity of the embeddings. This yields a significant improvement on WN18RR, as the descriptive information provides rich semantic cues. In contrast, the effect on FB15k-237 is marginal, since its descriptions contribute limited additional information beyond the already dense structural connections. 
In addition, the improvement on TransE is less pronounced than on other non-additive embedding models, which is consistent with the findings reported in PLM-Attack. This is likely due to the fact that Llama3 is a Transformer-based architecture that relies on scaled dot-product attention, it is inherently more difficult to transfer to additive models. 

\begin{figure}[!htbp]
    \centering
    \includegraphics[width=1\linewidth]{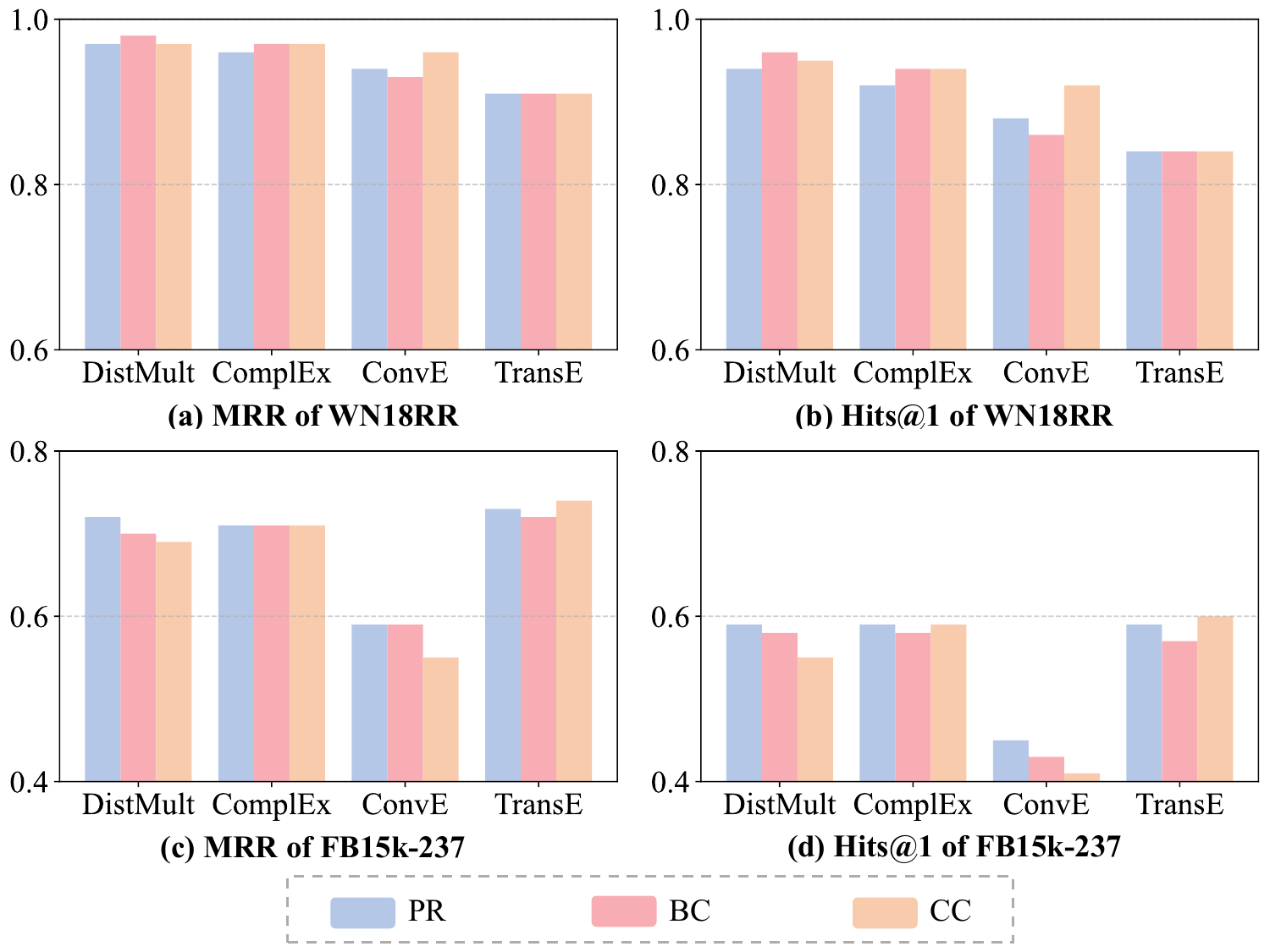}
    \caption{Ablation study on centrality-based filter. PR, BC, and CC denote pagerank, betweenness, and closeness centralities, respectively.}
    \label{fig:abl_centrality}
    \Description{}
\end{figure}

As shown in~\cref{fig:abl_centrality}, we analyze the diameter of the KG and select entities within 3 hops as the initial candidate set. Excessive hop distances reduce perturbation relevance, while insufficient hops limit misleading noise. For the two datasets, when selecting the top-30 entities for each centrality measure, the average candidate size is 46.16 and 57.58, respectively. This indicates that the overlap among the three filtered candidates is small. 
On WN18RR, the results across three centrality-based filters are similar, since inverse relations provide strong support. It is difficult to identify more adversarial entities $e_{noise}$ that could destabilize the influence of inverse relations.  

\section{Case Study}

\begin{table*}[!ht]
\centering
\caption{
Case study of a single triple attack on WN18RR with step-by-step reasoning. 
Green indicates accepted reason, and red denotes rejected reason. 
}
\label{tab:casestudy}
\input{tables/casestudy}

\end{table*}

As shown in~\cref{tab:casestudy,tab:casestudy_hesi}, we present a single triple attack case with step-by-step reasoning, demonstrating our framework's ability to generate human-readable explanations. 
It is noteworthy that the HoA-tuned LLM filters candidates more effectively than the semantics-based filter by leveraging structural information rather than mere lexical similarity, it filters the bidirectional evidence of the target triple as the Top1 candidate. 
The PR+BC+CC ensemble filters the candidate set containing multiple perturbative yet semantically proximate entities. 
For the deletion attack, the reasoning evaluates candidates one by one and selects candidate 1 because its inverse relation provides bidirectional corroboration with the target triple, the remaining candidates are rejected with explicit justifications, such as "not directly conflicting". For the addition attack, the reasoning prioritizes candidates with clear noun–adjective mappings, candidate 11, "unfitness," is semantically tied to "unwell" but misaligned with "feverish," thereby increasing confusability. 

\section{Conclusion}
In this paper, we propose LLMAtKGE, a unified framework leveraging LLMs to perform black-box attacks against knowledge graph embeddings KGEs, with the dual capability of target selection and human-readable explanation generation. It integrates the advantages of both parameter preservation and parameter updating, and further incorporates the semantics-based filter, centrality-based filters, and a HoA-tuning filter to reduce the candidate set and alleviate context overflow as well as reasoning hesitation, thereby enhancing attack effectiveness. We conduct extensive experiments on WN18RR and FB15k-237 across four KGEs, executing both deletion and addition attacks. The results show that, under black-box settings, our method consistently outperforms the strongest existing baselines. In many cases, it even approaches or surpasses the performance of white-box attacks, while also providing reasoning for the generated attacks. 


\clearpage
\balance
\bibliographystyle{ACM-Reference-Format}
\bibliography{reference}


\clearpage
\appendix 
\section{Prompt Template} \label{app:prompt_example}
We present the prompt template used for deletion attacks on WN18RR. For experiments on other datasets, the template remains identical except for the example component, which is replaced accordingly. Example instances are sampled from the validation set of KG, and the human-readable explanation may be produced by a fully-parameterized LLM such as ChatGPT-5. 
\begin{table}[!hb]
\centering

\input{tables/prompt_wn_del}

\end{table}

\section{Detailed Settings} \label{app:settings}
\subsection{Datasets}

\begin{table}[ht]
    \centering
    \caption{Summary statistics of datasets. }
    \label{tab:datasetset}

\input{tables/dataset_stat}
\end{table}

We conduct experimental analyses on two public datasets, as shown in~\cref{tab:datasetset}. WN18RR~\cite{WN18RR} is a subset of WordNet, containing 40,559 entities and 11 relation types. The triples primarily capture hierarchical relations among words, and the overall structure exhibits clear hierarchical organization and regularity. In contrast, FB15k-237~\cite{FB15k-237} is derived from Freebase, comprising 14,505 entities and 237 relation types. Its knowledge spans multiple domains such as movies, actors, awards, sports, and teams, with more diverse relation types and structural patterns. 

\subsection{Experimental Environment}
Qwen3-8B\footnote{\url{https://huggingface.co/Qwen/Qwen3-8B}} allows explicit control of the <thinking> field to switch chain-of-thought reasoning, thereby improving reasoning quality. Moreover, it supports a standardized output format for multiple-choice questions. Therefore, we adopt Qwen3-8B as the instruction LLM for reasoning.

For parameter-efficient fine-tuning, we employ LoRA with a rank of $r=64$, a scaling factor of $\alpha=16$, and a dropout rate of $0.05$. The LoRA adapters are applied to the query, key, value, and output projection layers within the attention blocks. We set the batch size to 16 and adopt the AdamW optimizer~\cite{loshchilov2019decoupled}. Under our experimental setup, the HoA tuning LLM trains for about 3 h on WN18RR and 8 h on FB15k-237. 

The implementation of LLMAtKGE is developed on PyTorch and Transformers, while the HoA filter is based on PEFT and NetworkX libraries. The computing environment is configured as follows: the OS is Ubuntu 22.04.5 LTS, with an Intel(R) Xeon(R) Platinum 8488C CPU, and five NVIDIA L20 48GB GPUs. 

\subsection{Hyperparameter Analysis}

\begin{table}[ht]
\centering
\caption{Performance under different TopK 
on FB15k-237.
}
\label{tab:hyper_del_fb}
\setlength{\tabcolsep}{6pt}
\renewcommand{\arraystretch}{1.2}

\input{tables/hyper_del_fb}

\end{table}

The choice of TopK significantly affects the performance of both the filters and the attacks. We  search multiple $K$ values. For the deletion attack, we set $K=\{3,5,10\}$ on WN18RR and $K=\{5,10,30\}$ on FB15k-237, as shown in~\cref{tab:hyper_del_fb}. For the addition attack, we experiment with $K=\{15,30,50\}$. 

\section{Case Study of Hesitation}
Due to limited space, we only present key reasoning traces. 
As shown in~\cref{tab:casestudy_hesi}, this LLM-based attempt often generates intermediate cues like "wait," reflecting transient hesitation or self-conflict within the reasoning process prior to final decision-making.  

\begin{table*}[!ht]
\centering
\caption{Case study of a deletion attack on WN18RR showing reasoning hesitation.}
\label{tab:casestudy_hesi}
\input{tables/casestudy_hesi}

\end{table*}

\end{document}

%% file: tables/main_result.tex
\newcommand{\ba}[1]{\textcolor{red}{\textbf{#1}}} 
\newcommand{\bb}[1]{\textbf{#1}} 
\newcommand{\sota}[1]{\underline{#1}} 

\newcommand{\ours}{\rowcolor[HTML]{E6F2FF}}
\newcommand{\white}{\cellcolor{white}}  

\begin{tabular}{@{}lll*{4}{cc}*{4}{cc}@{}}
\toprule
&&& \multicolumn{8}{c}{\textbf{WN18RR}} & \multicolumn{8}{c}{\textbf{FB15k-237}} \\
\cmidrule(lr){4-11}\cmidrule(l){12-19}
&&& \multicolumn{2}{c}{\textbf{DistMult}}
   & \multicolumn{2}{c}{\textbf{ComplEx}}
   & \multicolumn{2}{c}{\textbf{ConvE}}
   & \multicolumn{2}{c}{\textbf{TransE}}
   & \multicolumn{2}{c}{\textbf{DistMult}}
   & \multicolumn{2}{c}{\textbf{ComplEx}}
   & \multicolumn{2}{c}{\textbf{ConvE}}
   & \multicolumn{2}{c}{\textbf{TransE}} \\
&&& \textit{MRR} & \textit{H@1} & \textit{MRR} & \textit{H@1}
   & \textit{MRR} & \textit{H@1} & \textit{MRR} & \textit{H@1}
   & \textit{MRR} & \textit{H@1} & \textit{MRR} & \textit{H@1}
   & \textit{MRR} & \textit{H@1} & \textit{MRR} & \textit{H@1} \\
\midrule

\multirow{7}{*}{\rotatebox{90}{\textbf{Deletion}}}
  & \multirow{3}{*}{White-box} & Direct
  & 0.86 & 0.75 & 0.85 & 0.76 & 0.71 & 0.61 & 0.63 & 0.53
  & 0.63 & 0.50 & 0.60 & 0.43 & 0.61 & 0.48 & 0.63 & 0.47 \\
  &  & IF
  & 0.30 & 0.20 & 0.29 & 0.21 & 0.86 & 0.77 & 0.59 & 0.47
  & 0.62 & 0.48 & 0.61 & 0.45 & 0.53 & 0.35 & 0.63 & 0.46 \\
  &  & Random
  & 0.88 & 0.84 & 0.87 & 0.83 & 0.86 & 0.83 & 0.73 & 0.59
  & 0.66 & 0.54 & 0.68 & 0.53 & 0.62 & 0.47 & 0.76 & 0.65 \\
\cmidrule(l){2-19}
  & \multirow{3}{*}{Black-box} & AnyBURL
  & 0.29 & 0.20 & 0.31 & 0.22 & 0.22 & 0.17 & 0.54 & 0.44
  & 0.63 & 0.50 & 0.64 & 0.50 & 0.59 & 0.45 & 0.66 & 0.50 \\
  &  & PLM-Attack
  & 0.29 & 0.19 & 0.31 & 0.21 & 0.20 & 0.14 & 0.67 & 0.50
  & 0.57 & 0.43 & 0.60 & 0.43 & 0.56 & 0.38 & 0.66 & 0.51 \\
\ours   
  \white & \white & Ours
  & \ba{0.21} & \ba{0.14} & \ba{0.21} & \ba{0.14} & \ba{0.14} & \ba{0.11} & \sota{0.66} & \sota{0.50}
  & 0.65 & 0.48 & 0.63 & 0.48 & \ba{0.53} & \bb{0.36} & \bb{0.64} & \ba{0.46} \\
  
\bottomrule  

\multirow{7}{*}{\rotatebox{90}{\textbf{Addition}}}
  & \multirow{3}{*}{White-box} & Direct
  & 0.97 & 0.95 & 0.94 & 0.92 & 0.99 & 0.97 & 0.83 & 0.72
  & 0.64 & 0.53 & 0.69 & 0.55 & 0.65 & 0.52 & 0.73 & 0.62 \\
  &  & IF
  & 0.88 & 0.77 & 0.88 & 0.78 & 0.97 & 0.95 & 0.86 & 0.75
  & 0.67 & 0.55 & 0.70 & 0.59 & 0.63 & 0.50 & 0.75 & 0.64 \\
  &  & Random
  & 0.97 & 0.96 & 0.97 & 0.95 & 0.99 & 0.98 & 0.82 & 0.69
  & 0.66 & 0.54 & 0.69 & 0.57 & 0.66 & 0.53 & 0.74 & 0.63 \\
\cmidrule(l){2-19}
  & \multirow{3}{*}{Black-box} & AnyBURL
  & 0.86 & 0.74 & 0.82 & 0.69 & 0.88 & 0.77 & 0.81 & 0.66
  & 0.67 & 0.55 & 0.67 & 0.52 & 0.65 & 0.52 & 0.74 & 0.61 \\
  &  & PLM-Attack
  & 0.94 & 0.88 & 0.93 & 0.88 & 0.90 & 0.82 & 0.84 & 0.72
  & 0.65 & 0.51 & 0.66 & 0.53 & 0.63 & 0.45 & 0.74 & 0.62 \\
\ours   
  \white & \white & Ours
  & \sota{0.92} & \sota{0.84} & \sota{0.91} & \sota{0.81} & \ba{0.87} & \ba{0.75} & \ba{0.81} & \sota{0.68} 
  & 0.67 & \ba{0.51} & 0.68 & 0.54 & \ba{0.61} & 0.47 & \ba{0.65} & \ba{0.52} \\
\bottomrule
\end{tabular}

%% file: tables/abl_PE.tex
\begin{tabular}{@{}llcccccccc@{}}
\toprule
$\mathbf{E}$ & $\mathbf{R}$ 
& \multicolumn{2}{c}{\textbf{DistMult}} 
& \multicolumn{2}{c}{\textbf{ComplEx}} 
& \multicolumn{2}{c}{\textbf{ConvE}} 
& \multicolumn{2}{c}{\textbf{TransE}} \\ 
\cmidrule(l){3-10}
& & \textit{MRR} & \textit{H@1} 
  & \textit{MRR} & \textit{H@1} 
  & \textit{MRR} & \textit{H@1} 
  & \textit{MRR} & \textit{H@1} \\ 
\midrule
\xmark & \xmark & 0.24 & 0.17 & 0.25 & 0.19 & 0.17 & 0.13 & 0.69 & 0.53 \\
\xmark & \cmark & 0.32 & 0.23 & 0.34 & 0.28 & 0.24 & 0.20 & 0.70 & 0.53 \\
\cmark & \xmark & \textbf{0.21} & \textbf{0.14} 
                      & \textbf{0.21} & \textbf{0.14} 
                      & \textbf{0.14} & \textbf{0.11} 
                      & 0.73 & 0.61 \\
\cmark & \cmark & 0.22 & 0.14 & 0.23 & 0.18 & 0.16 & 0.12 & \textbf{0.66} & \textbf{0.50} \\ 
\bottomrule
\end{tabular}

%% file: tables/casestudy.tex
\newcommand{\acc}[1]{\textcolor[HTML]{2CA02C}{\textbf{#1}}}  
\newcommand{\error}[1]{\textcolor[HTML]{FF0000}{{#1}}}   
\newcommand{\answer}{\textcolor[HTML]{2CA02C}{\cmark \ }} 
\newcommand{\other}{\textcolor[HTML]{FF0000}{\xmark \ }} 
\newcommand{\lf}{\\ \vspace{-5pt}} 

\begin{tcolorbox}[
enhanced jigsaw,    
left=6pt, right=6pt, top=0pt, bottom=0pt, 
pad at break*=1mm,
colback=white!95!gray,
colframe=gray!50!black,
]
\small

\newcommand{\colw}{415pt}   
\setlength{\heavyrulewidth}{0.8pt}  
\setlength{\lightrulewidth}{0.4pt} 
\renewcommand{\arraystretch}{1.8}   
\newcolumntype{C}[1]{>{\centering\arraybackslash}m{#1}} 
\newcolumntype{L}[1]{>{\raggedright\arraybackslash}p{#1}} 
\begin{tabular}{C{0.1\linewidth} L{0.86\linewidth}}
\makecell[c]{target triple} & \makecell*[{{p{\colw}}}]{(feverish, derivationally related form, pyrexia)} \\
\toprule
\makecell[c]{semantics} & \makecell*[{{p{\colw}}}]{(pyrectic, derivationally related form, pyrexia), (pyrexia, derivationally related form, pyrectic), (pyrexia, derivationally related form, feverish), (hyperpyrexia, hypernym, pyrexia), (pyrexia, hypernym, symptom)} \\
\midrule
\makecell[c]{HoA tuning} & \makecell*[{{p{\colw}}}]{\answer \acc{(pyrexia, derivationally related form, feverish)}, (pyrectic, derivationally related form, pyrexia), (pyrexia, derivationally related form, pyrectic), (hyperpyrexia, hypernym, pyrexia), (pyrexia, hypernym, symptom)} \\
\toprule
\makecell[c]{PR} & \makecell*[{{p{\colw}}}]{medicine} \\
\midrule
\makecell[c]{BC} & \makecell*[{{p{\colw}}}]{set up} \\
\midrule
\makecell[c]{CC} & \makecell*[{{p{\colw}}}]{information} \\
\midrule
\makecell[c]{PR+BC+CC} & \makecell*[{{p{\colw}}}]{order, maturity, disorder, immaturity, unsoundness, obligation, noxious, end, hostility, conflict, \answer \acc{unfitness}, polyvalency, flaw, illumination, unwholesome} \\
\toprule
\makecell[c]{explanation\\of the\\deletion attack} & \makecell*[{{p{\colw}}}]{The target triple indicates that "feverish" is a derivationally related form of "pyrexia". So, in a knowledge graph, this would be a relation where one term is derived from another. 
Looking at the candidate triples: 
\lf
\answer1. (pyrexia, derivationally related form, feverish) – This is the inverse of the target triple. Since the target is (feverish, derivationally related form, pyrexia), this triple is (pyrexia, same relation, feverish). So they are \acc{bidirectional}. If the model has both, they reinforce each other. Removing one might weaken the model's confidence in the other. 
\lf
\other2. (pyrexia, derivationally related form, pyrectic) – This suggests that pyrexia and pyrectic are derivationally related. But pyrectic is an adjective meaning having a fever, so maybe it's related. But how does this affect the target triple? If the model has this, it might provide alternative relations for pyrexia, but \error{not directly conflicting with the target}. 
\lf
\other3. (pyrectic, derivationally related form, pyrexia) – This is the inverse of candidate 2. So similar to candidate 1, but with pyrectic instead of feverish. If the model has both 2 and 3, they might reinforce each other. But the target is about feverish and pyrexia. So removing this might \error{not directly affect} the target unless pyrectic is related to feverish. 
\lf
\other4. (hyperpyrexia, hypernym, pyrexia) – This is a hypernym relationship. Hyperpyrexia is a type of pyrexia (a high fever). So this is a different relation (hypernym) and not directly related to the derivational relation in the target. Removing this might \error{not affect} the target triple much. 
\lf
\other5. (pyrexia, hypernym, symptom) – This says pyrexia is a hypernym of symptom. But that's not correct. Symptoms are a category, and pyrexia is a specific symptom. So maybe this is incorrect. However, if the model has this, it might be a wrong relation. But how does that affect the target triple? The target is about derivational relations, so this hypernym relation might \error{not directly interfere}.
\lf
So for the current problem, the inverse of the target triple is \acc{candidate 1: (pyrexia, derivationally related form, feverish)}. If we remove this, the model would \acc{lose the bidirectional evidence for the derivational relation between feverish and pyrexia}. That would make the model less confident in predicting the target triple.} \\
\midrule
\makecell[c]{explanation\\of the\\addition attack} & \makecell*[{{p{\colw}}}]{The relation is “derivationally related form”. In linguistics, such forms are related through affixation or other morphological processes. The triple is (pyrexia, derivationally related form, feverish), so the model learns that “pyrexia” is derivationally related to “feverish”.
\lf
The replacement should be semantically close enough to “pyrexia” to appear in training, but different enough to disrupt the learned embeddings. We need a noun that can derive into an adjective like “feverish”.
Looking at the candidates:
1. order – noun; 
2. maturity – noun; 
3. disorder – noun; 
4. immaturity – noun; 
5. unsoundness – noun; 
6. obligation – noun; 
7. noxious – adjective; 
8. end – noun; 
9. hostility – noun; 
10. conflict – noun; 
11. unfitness – noun; 
12. polyvalency – noun; 
13. flaw – noun; 
14. illumination – noun; 
15. unwholesome – adjective; 
Nouns with clear adjective forms include: disorder→disordered, unfitness→unfit, flaw→flawed. “Feverish” is an adjective meaning having a fever; “pyrexia” is a medical noun for fever. Replacing “pyrexia” with a noun that derives to a different adjective should confuse the model.
\lf
\answer “Unfitness” (11) is a noun related to being unwell, whose adjective is “unfit”, not “feverish”; the model might \acc{expect “unfit” rather than “feverish”, disrupting the learned relation}. 
\lf
\other “Disorder” (3) is broader and would suggest “disordered”, which is \error{not specifically tied to feverish}. 
\lf
\other “Flaw” (13)→“flawed” is even \error{less related}.
\lf
Therefore, \acc{11 (unfitness)} is the best choice.}
\end{tabular}

\end{tcolorbox}

%% file: tables/prompt_wn_del.tex
\begin{tcolorbox}[
enhanced jigsaw,    
breakable,  
pad at break*=1mm,
colback=white!95!gray,
colframe=gray!50!black,
title={Prompt for deletion attack on WN18RR.}
]
\small
\begin{lstlisting}[
    breaklines=true,    % 自动换行
    xleftmargin=0pt, 
    breakindent=0pt, 
    columns=fullflexible    % 列对齐
]
You are an expert in knowledge graph reasoning and adversarial attack design. Your task is to select one candidate triple from the list below to delete, in order to destroy the ability of a knowledge graph embedding model (such as ConvE) to correctly predict the target triple. 
Example Target triple: (telephone, verb group, call). 
Example Candidate triples: 
1. (call, verb group, telephone)
2. (call in, hypernym, telephone)
3. (telephone, hypernym, telecommunicate)
4. (telephone, derivationally related form, telephony)
5. (telephony, derivationally related form, telephone). 
The target triple is (telephone verb group call). Among the candidates, triple 1 (call verb group telephone) is the inverse of the target and encodes the same semantic relation. Removing this mutual reinforcement (bidirectional "verb group") relation is likely to significantly degrade the model's confidence in predicting the target triple. It disrupts symmetric contextual evidence, which ConvE and similar embedding models heavily rely on. "answer": "3". 

Please process the following target triple ({target_triple}). 
Here are some candidate triples: {influence_triple_choice}. 
The details of entity is described as follows: {entity_desc}. 
Please show your choice in the answer field with only the choice number, e.g., "answer": "1". 
\end{lstlisting}
\end{tcolorbox}

%% file: tables/dataset_stat.tex
\begin{tabular}{@{}cccccc@{}}
\toprule
Dataset   & \#Ent & \#Rel & \#Train & \#Dev & \#Test \\ 
\midrule
WN18RR    & 40559 & 11    & 86835   & 2824  & 2924   \\
FB15k-237 & 14505 & 237   & 272115  & 17526 & 20438  \\ 
\bottomrule
\end{tabular}

%% file: tables/hyper_del_fb.tex
\begin{tabular}{@{}lcccccc@{}}
\toprule
\textbf{K} 
& \multicolumn{2}{c}{\textbf{DistMult}} 
& \multicolumn{2}{c}{\textbf{ComplEx}} 
& \multicolumn{2}{c}{\textbf{ConvE}} \\
\cmidrule(l){2-7}
& \textit{MRR} & \textit{H@1} 
& \textit{MRR} & \textit{H@1} 
& \textit{MRR} & \textit{H@1} \\
\midrule
5  & 0.70 & 0.54 & 0.70 & 0.53 & 0.60 & 0.39 \\
10 & 0.65 & \textbf{0.48} & \textbf{0.63} & \textbf{0.48} & 0.53 & \textbf{0.36} \\
30 & \textbf{0.64} & 0.50 & 0.65 & 0.52 & \textbf{0.52} & 0.37 \\
\bottomrule
\end{tabular}

%% file: tables/casestudy_hesi.tex
\newcommand{\acc}[1]{\textcolor[HTML]{2CA02C}{\textbf{#1}}}  
\newcommand{\error}[1]{\textcolor[HTML]{FF0000}{{#1}}}   
\newcommand{\answer}{\textcolor[HTML]{2CA02C}{\cmark \ }} 
\newcommand{\other}{\textcolor[HTML]{FF0000}{\xmark \ }} 
\newcommand{\lf}{\\ \vspace{-5pt}} 

\begin{tcolorbox}[
enhanced jigsaw,    
left=6pt, right=6pt, top=0pt, bottom=0pt, 
pad at break*=1mm,
colback=white!95!gray,
colframe=gray!50!black,
]
\small

\newcommand{\colw}{440pt}
\setlength{\heavyrulewidth}{0.8pt}  
\setlength{\lightrulewidth}{0.4pt} 
\renewcommand{\arraystretch}{1.8}   
\newcolumntype{C}[1]{>{\centering\arraybackslash}m{#1}} 
\newcolumntype{L}[1]{>{\raggedright\arraybackslash}p{#1}} 
\begin{tabular}{C{0.05\linewidth} L{0.9\linewidth}}
\makecell[c]{input} & \makecell*[{{p{\colw}}}]{
Please process the following target triple: \textbf{(throbbing derivationally related form throb)}. 
Here are some candidate triples: 
1. (throb derivationally related form throbbing)
2. (throbbing derivationally related form throb)
3. (throbbing derivationally related form thump)
} \\
\toprule
\makecell[c]{output} & \makecell*[{{p{\colw}}}]{
First, I should understand what a derivationally related form means. Dervationally related forms are words that are derived from each other, like how "throb" can become "throbbing" by adding a suffix. So, the target triple is saying that "throbbing" is a derivationally related form of "throb". 
Looking at the candidates:
\lf
1. (throb derivationally related form throbbing) – This seems to reverse the order of the target triple. The target is (A derivationally related form B), so this would be (B derivationally related form A). But since derivation can go both ways (like base form to derived form), this might still be correct. However, the original target is (throbbing derivationally related form throb), so the first candidate is the inverse. \error{But does that matter? }
\lf
2. (throbbing derivationally related form throb) – This is exactly the same as the target triple. So this would be the same, so semantically identical. \error{But wait}, the target is (throbbing derivationally related form throb), so candidate 2 is the same as the target. But the user is asking for the most semantically similar, so maybe the answer is 2? \error{But wait}, the example given in the problem had the target as (telephone verb group call) and the answer was (call verb group telephone) because it was the exact inverse. So in that case, even though the order was reversed, the relation was the same. So maybe in this case, candidate 1 is the inverse of the target, but the target is (A derivationally related form B), and candidate 1 is (B derivationally related form A). So the relation is the same, but the order is reversed. 
\lf
But the target is (throbbing derivationally related form throb), so the first element is the derived form, and the second is the base form. So candidate 2 is the same as the target, so it's exactly the same. \error{Wait}, but the target is (A, relation, B), and candidate 2 is (A, relation, B). So if the target is (throbbing, derivationally related form, throb), then candidate 2 is exactly the same. But the user's example had the target as (telephone verb group call), and the answer was (call verb group telephone) because it was the inverse. So in that example, the answer was the inverse, but the relation was the same. 
\lf
So, if the target is (A, relation, B), then the correct answer would be the one that has the same relation and the same elements, but possibly in reverse order. But in the example, the user considered the inverse as the correct answer. So perhaps in this case, the answer is candidate 1, \error{but wait}, the target is (A, relation, B), and candidate 1 is (B, relation, A). So the relation is the same. So the semantic relation is that the first element is a derivationally related form of the second. So if the target is (A, derivationally related form, B), then candidate 1 is (B, derivationally related form, A), which is the same relation but reversed. 
\lf
So, the key is to check if the relation is the same. In the target, the relation is "derivationally related form". So the first element is a derivationally related form of the second. So candidate 2 is (throbbing derivationally related form throb), which is exactly the same as the target. So candidate 2 is the same triple. But the user is asking to select the most semantically similar triple from the list. If candidate 2 is exactly the same as the target, then that's the most similar. However, maybe the user's example is different. 
\lf
So in this case, if candidate 2 is the same as the target, then that's the most similar. But why would the user have that as a candidate? \error{Wait}, the target is (throbbing derivationally related form throb), and candidate 2 is (throbbing derivationally related form throb), which is the same. So the answer would be 2. But the example in the problem shows that the inverse is considered the same. So perhaps the user is considering that the order doesn't matter, so both the target and candidate 1 are semantically similar. 
In this case, the relation is "derivationally related form". So if the target is (A, relation, B), then the inverse is (B, relation, A). So the relation is the same, and the words are the same. 
\lf
\error{
But I think the correct answer is 2, because it's the same as the target. 
So I'm torn. 
}
\lf
\error{
So the answer would be 1. 
But the target is (A, relation, B), and candidate 1 is (B, relation, A). 
So the answer is 1. 
}
\lf
\error{
But candidate 2 is the same as the target. 
So why is the answer not 2? 
}
\lf
\error{
I'm really confused now. I'm not sure. 
}
}
\end{tabular}

\end{tcolorbox}

%% file: main.bbl

\begin{thebibliography}{44}


\ifx \showCODEN    \undefined \def \showCODEN     #1{\unskip}     \fi
\ifx \showISBNx    \undefined \def \showISBNx     #1{\unskip}     \fi
\ifx \showISBNxiii \undefined \def \showISBNxiii  #1{\unskip}     \fi
\ifx \showISSN     \undefined \def \showISSN      #1{\unskip}     \fi
\ifx \showLCCN     \undefined \def \showLCCN      #1{\unskip}     \fi
\ifx \shownote     \undefined \def \shownote      #1{#1}          \fi
\ifx \showarticletitle \undefined \def \showarticletitle #1{#1}   \fi
\ifx \showURL      \undefined \def \showURL       {\relax}        \fi
\providecommand\bibfield[2]{#2}
\providecommand\bibinfo[2]{#2}
\providecommand\natexlab[1]{#1}
\providecommand\showeprint[2][]{arXiv:#2}

\bibitem[Betz et~al\mbox{.}(2022)]%
        {AnyBURL}
\bibfield{author}{\bibinfo{person}{Patrick Betz}, \bibinfo{person}{Christian Meilicke}, {and} \bibinfo{person}{Heiner Stuckenschmidt}.} \bibinfo{year}{2022}\natexlab{}.
\newblock \showarticletitle{{Adversarial Explanations for Knowledge Graph Embeddings}}. In \bibinfo{booktitle}{\emph{Proceedings of the Thirty-First International Joint Conference on Artificial Intelligence}}. \bibinfo{pages}{2820--2826}.
\newblock


\bibitem[Bhardwaj et~al\mbox{.}(2021a)]%
        {BaseAttack}
\bibfield{author}{\bibinfo{person}{Peru Bhardwaj}, \bibinfo{person}{John Kelleher}, \bibinfo{person}{Luca Costabello}, {and} \bibinfo{person}{Declan O'Sullivan}.} \bibinfo{year}{2021}\natexlab{a}.
\newblock \showarticletitle{{Adversarial Attacks on Knowledge Graph Embeddings via Instance Attribution Methods}}. In \bibinfo{booktitle}{\emph{Proceedings of the 2021 Conference on Empirical Methods in Natural Language Processing}}. \bibinfo{pages}{8225--8239}.
\newblock


\bibitem[Bhardwaj et~al\mbox{.}(2021b)]%
        {bhardwaj2021poisoning}
\bibfield{author}{\bibinfo{person}{Peru Bhardwaj}, \bibinfo{person}{John~D. Kelleher}, \bibinfo{person}{Luca Costabello}, {and} \bibinfo{person}{Declan O'Sullivan}.} \bibinfo{year}{2021}\natexlab{b}.
\newblock \showarticletitle{{Poisoning Knowledge Graph Embeddings via Relation Inference Patterns}}. In \bibinfo{booktitle}{\emph{Proceedings of the 59th Annual Meeting of the Association for Computational Linguistics and the 11th International Joint Conference on Natural Language Processing}}. \bibinfo{pages}{1875--1888}.
\newblock


\bibitem[Bordes et~al\mbox{.}(2013)]%
        {TransE}
\bibfield{author}{\bibinfo{person}{Antoine Bordes}, \bibinfo{person}{Nicolas Usunier}, \bibinfo{person}{Alberto {Garc{'{\i}}a-Dur{'a}n}}, \bibinfo{person}{Jason Weston}, {and} \bibinfo{person}{Oksana Yakhnenko}.} \bibinfo{year}{2013}\natexlab{}.
\newblock \showarticletitle{{Translating Embeddings for Modeling Multi-Relational Data}}. In \bibinfo{booktitle}{\emph{Advances in Neural Information Processing Systems}}. \bibinfo{pages}{2787--2795}.
\newblock


\bibitem[Brown et~al\mbox{.}(2020)]%
        {brown2020language}
\bibfield{author}{\bibinfo{person}{Tom Brown}, \bibinfo{person}{Benjamin Mann}, \bibinfo{person}{Nick Ryder}, \bibinfo{person}{Melanie Subbiah}, \bibinfo{person}{Jared~D. Kaplan}, \bibinfo{person}{Prafulla Dhariwal}, \bibinfo{person}{Arvind Neelakantan}, \bibinfo{person}{Pranav Shyam}, \bibinfo{person}{Girish Sastry}, \bibinfo{person}{Amanda Askell}, \bibinfo{person}{Sandhini Agarwal}, \bibinfo{person}{Ariel Herbert-Voss}, \bibinfo{person}{Gretchen Krueger}, \bibinfo{person}{Tom Henighan}, \bibinfo{person}{Rewon Child}, \bibinfo{person}{Aditya Ramesh}, \bibinfo{person}{Daniel~M. Ziegler}, \bibinfo{person}{Jeffrey Wu}, \bibinfo{person}{Clemens Winter}, \bibinfo{person}{Chris Hesse}, \bibinfo{person}{Mark Chen}, \bibinfo{person}{Eric Sigler}, \bibinfo{person}{Mateusz Litwin}, \bibinfo{person}{Scott Gray}, \bibinfo{person}{Benjamin Chess}, \bibinfo{person}{Jack Clark}, \bibinfo{person}{Christopher Berner}, \bibinfo{person}{Sam McCandlish}, \bibinfo{person}{Alec Radford}, \bibinfo{person}{Ilya Sutskever}, {and}
  \bibinfo{person}{Dario Amodei}.} \bibinfo{year}{2020}\natexlab{}.
\newblock \showarticletitle{{Language Models are Few-Shot Learners}}. In \bibinfo{booktitle}{\emph{Advances in Neural Information Processing Systems}}, Vol.~\bibinfo{volume}{33}.
\newblock


\bibitem[Chen et~al\mbox{.}(2025)]%
        {chen2025knowledge}
\bibfield{author}{\bibinfo{person}{Hanzhu Chen}, \bibinfo{person}{Xu Shen}, \bibinfo{person}{Jie Wang}, \bibinfo{person}{Zehao Wang}, \bibinfo{person}{Qitan Lv}, \bibinfo{person}{Junjie He}, \bibinfo{person}{Rong Wu}, \bibinfo{person}{Feng Wu}, {and} \bibinfo{person}{Jieping Ye}.} \bibinfo{year}{2025}\natexlab{}.
\newblock \showarticletitle{{Knowledge Graph Finetuning Enhances Knowledge Manipulation in Large Language Models}}. In \bibinfo{booktitle}{\emph{The Thirteenth International Conference on Learning Representations}}. \bibinfo{pages}{1--14}.
\newblock


\bibitem[Dettmers et~al\mbox{.}(2018)]%
        {WN18RR}
\bibfield{author}{\bibinfo{person}{Tim Dettmers}, \bibinfo{person}{Pasquale Minervini}, \bibinfo{person}{Pontus Stenetorp}, {and} \bibinfo{person}{Sebastian Riedel}.} \bibinfo{year}{2018}\natexlab{}.
\newblock \showarticletitle{{Convolutional 2D Knowledge Graph Embeddings}}.
\newblock \bibinfo{journal}{\emph{Proceedings of the AAAI Conference on Artificial Intelligence}} \bibinfo{volume}{32}, \bibinfo{number}{1} (\bibinfo{year}{2018}).
\newblock


\bibitem[Edge et~al\mbox{.}(2024)]%
        {edge2025local}
\bibfield{author}{\bibinfo{person}{Darren Edge}, \bibinfo{person}{Ha Trinh}, \bibinfo{person}{Newman Cheng}, \bibinfo{person}{Joshua Bradley}, \bibinfo{person}{Alex Chao}, \bibinfo{person}{Apurva Mody}, \bibinfo{person}{Steven Truitt}, \bibinfo{person}{Dasha Metropolitansky}, \bibinfo{person}{Robert~Osazuwa Ness}, {and} \bibinfo{person}{Jonathan Larson}.} \bibinfo{year}{2024}\natexlab{}.
\newblock \showarticletitle{{From Local to Global: A Graph RAG Approach to Query-Focused Summarization}}.
\newblock \bibinfo{journal}{\emph{arXiv preprint arXiv:2404.16130}} (\bibinfo{year}{2024}).
\newblock


\bibitem[Gerritse et~al\mbox{.}(2020)]%
        {gerritse2020graph}
\bibfield{author}{\bibinfo{person}{Emma~J Gerritse}, \bibinfo{person}{Faegheh Hasibi}, {and} \bibinfo{person}{Arjen~P de Vries}.} \bibinfo{year}{2020}\natexlab{}.
\newblock \showarticletitle{{Graph-Embedding Empowered Entity Retrieval}}. In \bibinfo{booktitle}{\emph{Advances in Information Retrieval: 42nd European Conference on IR Research, ECIR 2020, Lisbon, Portugal, April 14--17, 2020, Proceedings, Part I}}. \bibinfo{pages}{97--110}.
\newblock


\bibitem[Guo et~al\mbox{.}(2024a)]%
        {guo2024mkgl}
\bibfield{author}{\bibinfo{person}{Lingbing Guo}, \bibinfo{person}{Zhongpu Bo}, \bibinfo{person}{Zhuo Chen}, \bibinfo{person}{Yichi Zhang}, \bibinfo{person}{Jiaoyan Chen}, \bibinfo{person}{Yarong Lan}, \bibinfo{person}{Mengshu Sun}, \bibinfo{person}{Zhiqiang Zhang}, \bibinfo{person}{Yangyifei Luo}, \bibinfo{person}{Qian Li}, \bibinfo{person}{Qiang Zhang}, \bibinfo{person}{Wen Zhang}, {and} \bibinfo{person}{Huajun Chen}.} \bibinfo{year}{2024}\natexlab{a}.
\newblock \showarticletitle{{MKGL: Mastery of a Three-Word Language}}. In \bibinfo{booktitle}{\emph{Advances in Neural Information Processing Systems}}.
\newblock


\bibitem[Guo et~al\mbox{.}(2025)]%
        {guo2025kon}
\bibfield{author}{\bibinfo{person}{Lingbing Guo}, \bibinfo{person}{Yichi Zhang}, \bibinfo{person}{Zhongpu Bo}, \bibinfo{person}{Zhuo Chen}, \bibinfo{person}{Mengshu Sun}, \bibinfo{person}{Zhiqiang Zhang}, \bibinfo{person}{Wen Zhang}, {and} \bibinfo{person}{Huajun Chen}.} \bibinfo{year}{2025}\natexlab{}.
\newblock \showarticletitle{{K-ON: Stacking Knowledge On the Head Layer of Large Language Model}}. In \bibinfo{booktitle}{\emph{Proceedings of the AAAI Conference on Artificial Intelligence}}, Vol.~\bibinfo{volume}{39}. \bibinfo{pages}{11745--11753}.
\newblock


\bibitem[Guo et~al\mbox{.}(2024b)]%
        {guo2024lightrag}
\bibfield{author}{\bibinfo{person}{Zirui Guo}, \bibinfo{person}{Lianghao Xia}, \bibinfo{person}{Yanhua Yu}, \bibinfo{person}{Tu Ao}, {and} \bibinfo{person}{Chao Huang}.} \bibinfo{year}{2024}\natexlab{b}.
\newblock \showarticletitle{{LightRAG: Simple and Fast Retrieval-Augmented Generation}}.
\newblock \bibinfo{journal}{\emph{arXiv preprint arXiv:2410.05779}} (\bibinfo{year}{2024}).
\newblock


\bibitem[Gutierrez et~al\mbox{.}(2024)]%
        {gutierrez2024hipporag}
\bibfield{author}{\bibinfo{person}{Bernal~Jimenez Gutierrez}, \bibinfo{person}{Yiheng Shu}, \bibinfo{person}{Yu Gu}, \bibinfo{person}{Michihiro Yasunaga}, {and} \bibinfo{person}{Yu Su}.} \bibinfo{year}{2024}\natexlab{}.
\newblock \showarticletitle{{HippoRAG: Neurobiologically Inspired Long-Term Memory for Large Language Models}}. In \bibinfo{booktitle}{\emph{Advances in Neural Information Processing Systems}}.
\newblock


\bibitem[Guti{'e}rrez et~al\mbox{.}(2025)]%
        {gutierrez2025rag}
\bibfield{author}{\bibinfo{person}{Bernal~Jim{'e}nez Guti{'e}rrez}, \bibinfo{person}{Yiheng Shu}, \bibinfo{person}{Weijian Qi}, \bibinfo{person}{Sizhe Zhou}, {and} \bibinfo{person}{Yu Su}.} \bibinfo{year}{2025}\natexlab{}.
\newblock \showarticletitle{{From RAG to Memory: Non-Parametric Continual Learning for Large Language Models}}.
\newblock \bibinfo{journal}{\emph{arXiv preprint arXiv:2502.14802}} (\bibinfo{year}{2025}).
\newblock


\bibitem[Hamilton et~al\mbox{.}(2017)]%
        {hamilton2017inductive}
\bibfield{author}{\bibinfo{person}{William~L. Hamilton}, \bibinfo{person}{Zhitao Ying}, {and} \bibinfo{person}{Jure Leskovec}.} \bibinfo{year}{2017}\natexlab{}.
\newblock \showarticletitle{{Inductive Representation Learning on Large Graphs}}. In \bibinfo{booktitle}{\emph{Advances in Neural Information Processing Systems}}. \bibinfo{pages}{1024--1034}.
\newblock


\bibitem[Huang et~al\mbox{.}(2019)]%
        {huang2019knowledge}
\bibfield{author}{\bibinfo{person}{Xiao Huang}, \bibinfo{person}{Jingyuan Zhang}, \bibinfo{person}{Dingcheng Li}, {and} \bibinfo{person}{Ping Li}.} \bibinfo{year}{2019}\natexlab{}.
\newblock \showarticletitle{{Knowledge Graph Embedding Based Question Answering}}. In \bibinfo{booktitle}{\emph{Proceedings of the Twelfth ACM International Conference on Web Search and Data Mining}}. \bibinfo{pages}{105--113}.
\newblock


\bibitem[Kapoor et~al\mbox{.}(2025)]%
        {kapoor2025robustness}
\bibfield{author}{\bibinfo{person}{Sourabh Kapoor}, \bibinfo{person}{Arnab Sharma}, \bibinfo{person}{Michael R{"o}der}, \bibinfo{person}{Caglar Demir}, {and} \bibinfo{person}{Axel-Cyrille Ngonga~Ngomo}.} \bibinfo{year}{2025}\natexlab{}.
\newblock \showarticletitle{{Robustness Evaluation of Knowledge Graph Embedding Models Under Non-Targeted Attacks}}. In \bibinfo{booktitle}{\emph{The Semantic Web}}, Vol.~\bibinfo{volume}{15718}. \bibinfo{pages}{264--281}.
\newblock


\bibitem[Kipf and Welling(2017)]%
        {kipf2017semisupervised}
\bibfield{author}{\bibinfo{person}{Thomas~N. Kipf} {and} \bibinfo{person}{Max Welling}.} \bibinfo{year}{2017}\natexlab{}.
\newblock \showarticletitle{{Semi-Supervised Classification with Graph Convolutional Networks}}. In \bibinfo{booktitle}{\emph{The Fifth International Conference on Learning Representations}}. \bibinfo{pages}{1--14}.
\newblock


\bibitem[Liu et~al\mbox{.}(2025)]%
        {liu2025dual}
\bibfield{author}{\bibinfo{person}{Guangyi Liu}, \bibinfo{person}{Yongqi Zhang}, \bibinfo{person}{Yong Li}, {and} \bibinfo{person}{Quanming Yao}.} \bibinfo{year}{2025}\natexlab{}.
\newblock \showarticletitle{{Dual Reasoning: A GNN-LLM Collaborative Framework for Knowledge Graph Question Answering}}. In \bibinfo{booktitle}{\emph{Conference on Parsimony and Learning}}, Vol.~\bibinfo{volume}{280}. \bibinfo{pages}{351--372}.
\newblock


\bibitem[Loshchilov and Hutter(2019)]%
        {loshchilov2019decoupled}
\bibfield{author}{\bibinfo{person}{Ilya Loshchilov} {and} \bibinfo{person}{Frank Hutter}.} \bibinfo{year}{2019}\natexlab{}.
\newblock \showarticletitle{{Decoupled Weight Decay Regularization}}. In \bibinfo{booktitle}{\emph{Proceedings of the 7th International Conference on Learning Representations}}.
\newblock


\bibitem[Luo et~al\mbox{.}(2024)]%
        {luo2024reasoning}
\bibfield{author}{\bibinfo{person}{Linhao Luo}, \bibinfo{person}{Yuan-Fang Li}, \bibinfo{person}{Gholamreza Haffari}, {and} \bibinfo{person}{Shirui Pan}.} \bibinfo{year}{2024}\natexlab{}.
\newblock \showarticletitle{{Reasoning on Graphs: Faithful and Interpretable Large Language Model Reasoning}}. In \bibinfo{booktitle}{\emph{The Twelfth International Conference on Learning Representations}}. \bibinfo{pages}{1--14}.
\newblock


\bibitem[Luo et~al\mbox{.}(2025)]%
        {luo2025graphconstrained}
\bibfield{author}{\bibinfo{person}{Linhao Luo}, \bibinfo{person}{Zicheng Zhao}, \bibinfo{person}{Gholamreza Haffari}, \bibinfo{person}{Yuan-Fang Li}, \bibinfo{person}{Chen Gong}, {and} \bibinfo{person}{Shirui Pan}.} \bibinfo{year}{2025}\natexlab{}.
\newblock \showarticletitle{{Graph-Constrained Reasoning: Faithful Reasoning on Knowledge Graphs with Large Language Models}}.
\newblock \bibinfo{journal}{\emph{arXiv preprint arXiv:2410.13080}} (\bibinfo{year}{2025}).
\newblock


\bibitem[Ma et~al\mbox{.}(2025)]%
        {ma2025thinkongraph}
\bibfield{author}{\bibinfo{person}{Shengjie Ma}, \bibinfo{person}{Chengjin Xu}, \bibinfo{person}{Xuhui Jiang}, \bibinfo{person}{Muzhi Li}, \bibinfo{person}{Huaren Qu}, \bibinfo{person}{Cehao Yang}, \bibinfo{person}{Jiaxin Mao}, {and} \bibinfo{person}{Jian Guo}.} \bibinfo{year}{2025}\natexlab{}.
\newblock \showarticletitle{{Think-on-Graph 2.0: Deep and Faithful Large Language Model Reasoning with Knowledge-Guided Retrieval Augmented Generation}}. In \bibinfo{booktitle}{\emph{The Thirteenth International Conference on Learning Representations}}. \bibinfo{pages}{1--14}.
\newblock


\bibitem[Nikolaev and Kotov(2020)]%
        {nikolaev2020joint}
\bibfield{author}{\bibinfo{person}{Fedor Nikolaev} {and} \bibinfo{person}{Alexander Kotov}.} \bibinfo{year}{2020}\natexlab{}.
\newblock \showarticletitle{{Joint Word and Entity Embeddings for Entity Retrieval from a Knowledge Graph}}. In \bibinfo{booktitle}{\emph{Advances in Information Retrieval: 42nd European Conference on IR Research, ECIR 2020, Lisbon, Portugal, April 14--17, 2020, Proceedings, Part I}}. \bibinfo{pages}{141--155}.
\newblock


\bibitem[Pezeshkpour et~al\mbox{.}(2019)]%
        {CRIAGE}
\bibfield{author}{\bibinfo{person}{Pouya Pezeshkpour}, \bibinfo{person}{Yifan Tian}, {and} \bibinfo{person}{Sameer Singh}.} \bibinfo{year}{2019}\natexlab{}.
\newblock \showarticletitle{{Investigating Robustness and Interpretability of Link Prediction via Adversarial Modifications}}. In \bibinfo{booktitle}{\emph{Proceedings of the 2019 Conference of the North American Chapter of the Association for Computational Linguistics: Human Language Technologies}}. \bibinfo{pages}{3336--3347}.
\newblock


\bibitem[Sarthi et~al\mbox{.}(2024)]%
        {sarthi2024raptor}
\bibfield{author}{\bibinfo{person}{Parth Sarthi}, \bibinfo{person}{Salman Abdullah}, \bibinfo{person}{Aditi Tuli}, \bibinfo{person}{Shubh Khanna}, \bibinfo{person}{Anna Goldie}, {and} \bibinfo{person}{Christopher~D. Manning}.} \bibinfo{year}{2024}\natexlab{}.
\newblock \showarticletitle{{RAPTOR: Recursive Abstractive Processing for Tree-Organized Retrieval}}. In \bibinfo{booktitle}{\emph{The Twelfth International Conference on Learning Representations}}. \bibinfo{pages}{1--14}.
\newblock


\bibitem[Saxena et~al\mbox{.}(2020)]%
        {saxena2020improving}
\bibfield{author}{\bibinfo{person}{Apoorv Saxena}, \bibinfo{person}{Aditay Tripathi}, {and} \bibinfo{person}{Partha Talukdar}.} \bibinfo{year}{2020}\natexlab{}.
\newblock \showarticletitle{{Improving Multi-Hop Question Answering over Knowledge Graphs Using Knowledge Base Embeddings}}. In \bibinfo{booktitle}{\emph{Proceedings of the 58th Annual Meeting of the Association for Computational Linguistics}}. \bibinfo{pages}{4498--4507}.
\newblock


\bibitem[Sun et~al\mbox{.}(2024)]%
        {sun2024thinkongraph}
\bibfield{author}{\bibinfo{person}{Jiashuo Sun}, \bibinfo{person}{Chengjin Xu}, \bibinfo{person}{Lumingyuan Tang}, \bibinfo{person}{Saizhuo Wang}, \bibinfo{person}{Chen Lin}, \bibinfo{person}{Yeyun Gong}, \bibinfo{person}{Lionel~M. Ni}, \bibinfo{person}{Heung-Yeung Shum}, {and} \bibinfo{person}{Jian Guo}.} \bibinfo{year}{2024}\natexlab{}.
\newblock \showarticletitle{{Think-on-Graph: Deep and Responsible Reasoning of Large Language Model on Knowledge Graph}}. In \bibinfo{booktitle}{\emph{The Twelfth International Conference on Learning Representations}}. \bibinfo{pages}{1--14}.
\newblock


\bibitem[Sun et~al\mbox{.}(2019)]%
        {sun2019rotate}
\bibfield{author}{\bibinfo{person}{Zhiqing Sun}, \bibinfo{person}{Zhi-Hong Deng}, \bibinfo{person}{Jian-Yun Nie}, {and} \bibinfo{person}{Jian Tang}.} \bibinfo{year}{2019}\natexlab{}.
\newblock \showarticletitle{{RotatE: Knowledge Graph Embedding by Relational Rotation in Complex Space}}. In \bibinfo{booktitle}{\emph{Proceedings of the 7th International Conference on Learning Representations}}. \bibinfo{pages}{1--14}.
\newblock


\bibitem[Toutanova et~al\mbox{.}(2015)]%
        {FB15k-237}
\bibfield{author}{\bibinfo{person}{Kristina Toutanova}, \bibinfo{person}{Danqi Chen}, \bibinfo{person}{Patrick Pantel}, \bibinfo{person}{Hoifung Poon}, \bibinfo{person}{Pallavi Choudhury}, {and} \bibinfo{person}{Michael Gamon}.} \bibinfo{year}{2015}\natexlab{}.
\newblock \showarticletitle{{Representing Text for Joint Embedding of Text and Knowledge Bases}}. In \bibinfo{booktitle}{\emph{Proceedings of the 2015 Conference on Empirical Methods in Natural Language Processing}}. \bibinfo{pages}{1499--1509}.
\newblock


\bibitem[Trouillon et~al\mbox{.}(2016)]%
        {ComplEx}
\bibfield{author}{\bibinfo{person}{Th{'e}o Trouillon}, \bibinfo{person}{Johannes Welbl}, \bibinfo{person}{Sebastian Riedel}, \bibinfo{person}{Eric Gaussier}, {and} \bibinfo{person}{Guillaume Bouchard}.} \bibinfo{year}{2016}\natexlab{}.
\newblock \showarticletitle{{Complex Embeddings for Simple Link Prediction}}. In \bibinfo{booktitle}{\emph{Proceedings of The 33rd International Conference on Machine Learning}}. \bibinfo{pages}{2071--2080}.
\newblock


\bibitem[Velickovic et~al\mbox{.}(2018)]%
        {velickovic2018graph}
\bibfield{author}{\bibinfo{person}{Petar Velickovic}, \bibinfo{person}{Guillem Cucurull}, \bibinfo{person}{Arantxa Casanova}, \bibinfo{person}{Adriana Romero}, \bibinfo{person}{Pietro Li{`o}}, {and} \bibinfo{person}{Yoshua Bengio}.} \bibinfo{year}{2018}\natexlab{}.
\newblock \showarticletitle{{Graph Attention Networks}}. In \bibinfo{booktitle}{\emph{The Sixth International Conference on Learning Representations}}. \bibinfo{pages}{1--12}.
\newblock


\bibitem[Wang et~al\mbox{.}(2024)]%
        {wang2024learning}
\bibfield{author}{\bibinfo{person}{Junjie Wang}, \bibinfo{person}{Mingyang Chen}, \bibinfo{person}{Binbin Hu}, \bibinfo{person}{Dan Yang}, \bibinfo{person}{Ziqi Liu}, \bibinfo{person}{Yue Shen}, \bibinfo{person}{Peng Wei}, \bibinfo{person}{Zhiqiang Zhang}, \bibinfo{person}{Jinjie Gu}, \bibinfo{person}{Jun Zhou}, \bibinfo{person}{Jeff~Z. Pan}, \bibinfo{person}{Wen Zhang}, {and} \bibinfo{person}{Huajun Chen}.} \bibinfo{year}{2024}\natexlab{}.
\newblock \showarticletitle{Learning to Plan for Retrieval-Augmented Large Language Models from Knowledge Graphs}. In \bibinfo{booktitle}{\emph{Findings of the Association for Computational Linguistics: EMNLP 2024, Miami, Florida, USA, November 12-16, 2024}}. \bibinfo{pages}{7813--7835}.
\newblock


\bibitem[Wei et~al\mbox{.}(2023)]%
        {wei2023kicgpt}
\bibfield{author}{\bibinfo{person}{Yanbin Wei}, \bibinfo{person}{Qiushi Huang}, \bibinfo{person}{James~T. Kwok}, {and} \bibinfo{person}{Yu Zhang}.} \bibinfo{year}{2023}\natexlab{}.
\newblock \showarticletitle{{KICGPT: Large Language Model with Knowledge in Context for Knowledge Graph Completion}}. In \bibinfo{booktitle}{\emph{Findings of the Association for Computational Linguistics: EMNLP 2023}}. \bibinfo{pages}{8667--8683}.
\newblock


\bibitem[Wu et~al\mbox{.}(2019)]%
        {wu2019simplifying}
\bibfield{author}{\bibinfo{person}{Felix Wu}, \bibinfo{person}{Amauri Souza}, \bibinfo{person}{Tianyi Zhang}, \bibinfo{person}{Christopher Fifty}, \bibinfo{person}{Tao Yu}, {and} \bibinfo{person}{Kilian Weinberger}.} \bibinfo{year}{2019}\natexlab{}.
\newblock \showarticletitle{{Simplifying Graph Convolutional Networks}}. In \bibinfo{booktitle}{\emph{Proceedings of the 36th International Conference on Machine Learning}}, Vol.~\bibinfo{volume}{97}. \bibinfo{pages}{6861--6871}.
\newblock


\bibitem[Xu et~al\mbox{.}(2024)]%
        {xu2024multiperspective}
\bibfield{author}{\bibinfo{person}{Derong Xu}, \bibinfo{person}{Ziheng Zhang}, \bibinfo{person}{Zhenxi Lin}, \bibinfo{person}{Xian Wu}, \bibinfo{person}{Zhihong Zhu}, \bibinfo{person}{Tong Xu}, \bibinfo{person}{Xiangyu Zhao}, \bibinfo{person}{Yefeng Zheng}, {and} \bibinfo{person}{Enhong Chen}.} \bibinfo{year}{2024}\natexlab{}.
\newblock \showarticletitle{{Multi-Perspective Improvement of Knowledge Graph Completion with Large Language Models}}.
\newblock \bibinfo{journal}{\emph{arXiv preprint arXiv:2403.01972}} (\bibinfo{year}{2024}).
\newblock


\bibitem[Yang et~al\mbox{.}(2015)]%
        {DistMult}
\bibfield{author}{\bibinfo{person}{Bishan Yang}, \bibinfo{person}{Wen-tau Yih}, \bibinfo{person}{Xiaodong He}, \bibinfo{person}{Jianfeng Gao}, {and} \bibinfo{person}{Li Deng}.} \bibinfo{year}{2015}\natexlab{}.
\newblock \showarticletitle{{Embedding Entities and Relations for Learning and Inference in Knowledge Bases}}.
\newblock \bibinfo{journal}{\emph{arXiv preprint arXiv:1412.6575}} (\bibinfo{year}{2015}).
\newblock


\bibitem[Yang et~al\mbox{.}(2024)]%
        {PLM-Attack}
\bibfield{author}{\bibinfo{person}{Guangqian Yang}, \bibinfo{person}{Lei Zhang}, \bibinfo{person}{Yi Liu}, \bibinfo{person}{Hongtao Xie}, {and} \bibinfo{person}{Zhendong Mao}.} \bibinfo{year}{2024}\natexlab{}.
\newblock \showarticletitle{{Exploiting Pre-Trained Language Models for Black-Box Attack against Knowledge Graph Embeddings}}.
\newblock \bibinfo{journal}{\emph{ACM Transactions on Knowledge Discovery from Data}} \bibinfo{volume}{19}, \bibinfo{number}{1} (\bibinfo{year}{2024}), \bibinfo{pages}{1--14}.
\newblock


\bibitem[Yao et~al\mbox{.}(2025)]%
        {yao2025exploring}
\bibfield{author}{\bibinfo{person}{Liang Yao}, \bibinfo{person}{Jiazhen Peng}, \bibinfo{person}{Chengsheng Mao}, {and} \bibinfo{person}{Yuan Luo}.} \bibinfo{year}{2025}\natexlab{}.
\newblock \showarticletitle{{Exploring Large Language Models for Knowledge Graph Completion}}. In \bibinfo{booktitle}{\emph{2025 IEEE International Conference on Acoustics, Speech and Signal Processing}}. \bibinfo{pages}{1--5}.
\newblock


\bibitem[Zhang et~al\mbox{.}(2019)]%
        {zhang2019data}
\bibfield{author}{\bibinfo{person}{Hengtong Zhang}, \bibinfo{person}{Tianhang Zheng}, \bibinfo{person}{Jing Gao}, \bibinfo{person}{Chenglin Miao}, \bibinfo{person}{Lu Su}, \bibinfo{person}{Yaliang Li}, {and} \bibinfo{person}{Kui Ren}.} \bibinfo{year}{2019}\natexlab{}.
\newblock \showarticletitle{{Data Poisoning Attack against Knowledge Graph Embedding}}. In \bibinfo{booktitle}{\emph{Proceedings of the Twenty-Eighth International Joint Conference on Artificial Intelligence}}. \bibinfo{pages}{4853--4859}.
\newblock


\bibitem[Zhang et~al\mbox{.}(2024)]%
        {KoPA}
\bibfield{author}{\bibinfo{person}{Yichi Zhang}, \bibinfo{person}{Zhuo Chen}, \bibinfo{person}{Lingbing Guo}, \bibinfo{person}{Yajing Xu}, \bibinfo{person}{Wen Zhang}, {and} \bibinfo{person}{Huajun Chen}.} \bibinfo{year}{2024}\natexlab{}.
\newblock \showarticletitle{{Making Large Language Models Perform Better in Knowledge Graph Completion}}. In \bibinfo{booktitle}{\emph{Proceedings of the 32nd ACM International Conference on Multimedia}}. \bibinfo{pages}{233--242}.
\newblock


\bibitem[Zhang et~al\mbox{.}(2025)]%
        {zhang2025qwen3}
\bibfield{author}{\bibinfo{person}{Yanzhao Zhang}, \bibinfo{person}{Mingxin Li}, \bibinfo{person}{Dingkun Long}, \bibinfo{person}{Xin Zhang}, \bibinfo{person}{Huan Lin}, \bibinfo{person}{Baosong Yang}, \bibinfo{person}{Pengjun Xie}, \bibinfo{person}{An Yang}, \bibinfo{person}{Dayiheng Liu}, \bibinfo{person}{Junyang Lin}, \bibinfo{person}{Fei Huang}, {and} \bibinfo{person}{Jingren Zhou}.} \bibinfo{year}{2025}\natexlab{}.
\newblock \showarticletitle{{Qwen3 Embedding: Advancing Text Embedding and Reranking through Foundation Models}}.
\newblock \bibinfo{journal}{\emph{arXiv preprint arXiv:2506.05176}} (\bibinfo{year}{2025}).
\newblock


\bibitem[Zhao et~al\mbox{.}(2024)]%
        {zhao2024untargeted}
\bibfield{author}{\bibinfo{person}{Tianzhe Zhao}, \bibinfo{person}{Jiaoyan Chen}, \bibinfo{person}{Yanchi Ru}, \bibinfo{person}{Qika Lin}, \bibinfo{person}{Yuxia Geng}, {and} \bibinfo{person}{Jun Liu}.} \bibinfo{year}{2024}\natexlab{}.
\newblock \showarticletitle{{Untargeted Adversarial Attack on Knowledge Graph Embeddings}}. In \bibinfo{booktitle}{\emph{Proceedings of the 47th International ACM SIGIR Conference on Research and Development in Information Retrieval}}. \bibinfo{pages}{1701--1711}.
\newblock


\bibitem[Zhou et~al\mbox{.}(2024)]%
        {zhou2024poisoning}
\bibfield{author}{\bibinfo{person}{Enyuan Zhou}, \bibinfo{person}{Song Guo}, \bibinfo{person}{Zhixiu Ma}, \bibinfo{person}{Zicong Hong}, \bibinfo{person}{Tao Guo}, {and} \bibinfo{person}{Peiran Dong}.} \bibinfo{year}{2024}\natexlab{}.
\newblock \showarticletitle{{Poisoning Attack on Federated Knowledge Graph Embedding}}. In \bibinfo{booktitle}{\emph{Proceedings of the ACM Web Conference 2024}}. \bibinfo{pages}{1998--2008}.
\newblock


\end{thebibliography}
